\documentclass[10pt,twocolumn,letterpaper]{article}

\usepackage[pagenumbers]{iccv} 

\definecolor{iccvblue}{rgb}{0.21,0.49,0.74}
\definecolor{unsipervised}{RGB}{213, 232, 212}
\definecolor{supervised}{RGB}{200, 200, 225}
\definecolor{paired}{RGB}{248, 206, 204}
\definecolor{best}{HTML}{fdf4bb}

\usepackage[pagebackref,breaklinks,colorlinks,allcolors=iccvblue]{hyperref}
\usepackage[T1]{fontenc}
\def\paperID{8451} 
\def\confName{ICCV}
\def\confYear{2025}

\title{Color Matching Using Hypernetwork-Based Kolmogorov-Arnold Networks}

\author{Artem Nikonorov$^{1}$$^{*}$ \quad Georgy Perevozchikov$^{2}$$^{*}$ \quad Andrei Korepanov$^{1}$ \quad Nancy Mehta$^{2}$$^{**}$ \\ \quad Mahmoud Afifi$^3$$^{\dagger}$ \quad Egor Ershov$^{4,5,6}$ \quad Radu Timofte$^{2}$ \vspace{0.3em} \\
{\normalsize $^1$Samara National Research University} \quad
{\normalsize $^2$Computer Vision Lab, CAIDAS \& IFI, University of W\"urzburg} \quad \\
{\normalsize $^3$York University} \quad
{\normalsize $^4$Institute for Information Transmission Problems RAS} \quad \\
{\normalsize $^5$Moscow Institute of Physics and Technologies} \quad
{\normalsize $^6$Artificial Intelligence Research Institute}
}

\begin{document}
\maketitle
\def\thefootnote{*}\footnotetext{These Authors Contributed Equally}
\def\thefootnote{**}\footnotetext{Corresponding Author}
\def\thefootnote{$\dagger$}\footnotetext{Now at Samsung}
\begin{abstract}
We present \emph{cmKAN}, a versatile framework for color matching. 
Given an input image with colors from a source color distribution, our method effectively and accurately maps these colors to match a target color distribution in both supervised and unsupervised settings. 
Our framework leverages the spline capabilities of Kolmogorov-Arnold Networks (KANs) to model the color matching between source and target distributions. 
Specifically, we developed a hypernetwork that generates spatially varying weight maps to control the nonlinear splines of a KAN, enabling accurate color matching. 
As part of this work, we introduce a first large-scale dataset of paired images captured by two distinct cameras and evaluate the efficacy of our and existing methods in matching colors. 
We evaluated our approach across various color-matching tasks, including: (1) \emph{raw-to-raw mapping}, where the source color distribution is in one camera’s raw color space and the target in another camera’s raw space; (2) \emph{raw-to-sRGB mapping}, where the source color distribution is in a camera’s raw space and the target is in the display sRGB space, emulating the color rendering of a camera ISP; and (3) \emph{sRGB-to-sRGB mapping}, where the goal is to transfer colors from a source sRGB space (e.g., produced by a source camera ISP) to a target sRGB space (e.g., from a different camera ISP). 
The results show that our method outperforms existing approaches by $37.3\%$ on average for supervised and unsupervised cases while remaining lightweight compared to other methods.
The codes, dataset, and pre-trained models are available at: \url{https://github.com/gosha20777/cmKAN.git} 

\end{abstract}
\begin{figure}[!t]
  \centering
\includegraphics[width=0.9\linewidth]{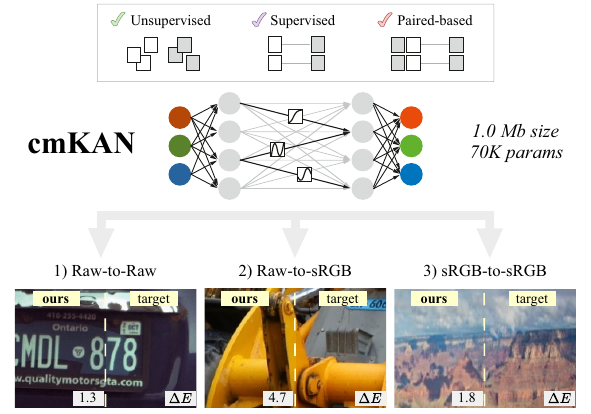}
\vspace{-2mm}
  \caption{We present \emph{cmKAN}, a learning framework for color matching. Our framework is versatile, supporting three color-matching scenarios: supervised and unsupervised offline training, as well as paired-based optimization. It is suitable for different color-matching tasks, such as (1) \emph{raw-to-raw}, (2) \emph{raw-to-sRGB} and (3) \emph{sRGB-to-sRGB} mapping. Compared to other methods (\eg, SIRLUT~\cite{li2024sirlut}), our method demonstrates promising results across all tasks with a small number of parameters.}
  \vspace{-4mm}
  \label{fig:teaser}
\end{figure}
\section{Introduction and Related Work}
\vspace{-1mm}
\label{sec:intro}

Camera Image Signal Processors (ISPs) consist of various modules that process input image colors to produce the final output \cite{nakamura2017image, heide2014flexisp, karaimer2016software, delbracio2021mobile, brown2023color}. 
These ISPs perform spatially nonuniform color transformations that map the initial image colors from the camera's raw space -- defined by the camera's characteristics -- to standardized color spaces (\eg, sRGB, DCI P3, \textit{etc}) \cite{nakamura2017image, brown2023color}. 

However, variations in the final rendered colors arise when the same scene is captured by different camera systems. These discrepancies arises from several factors, including the initial raw colors produced by the camera sensor and the diverse algorithms employed by ISPs, which can vary in accuracy and reflect the desired styles that the camera manufacturer aims for in the final image \cite{zhou2007image, nguyen2014raw, delbracio2021mobile, tseng2022neural}.  

Color matching, or more specifically color stabilization, aims to adjust the colors of an input image (source) to closely resemble the colors of a reference image (target) that captures the same scene \cite{AN1, AN2}. Typically, these images are taken by different cameras or by the same camera but in different color spaces \cite{AN2, finlayson2015color}. Color matching can thus be considered a subset of color transfer, focused specifically on aligning colors between source and target images. In contrast, color transfer generally involves images from different scenes \cite{faridul2014survey, lv2024color} and seeks to transfer the ``feel'' and ``look'' of the target image to the source image \cite{nguyen2014illuminant}, often resulting in extensive image recoloring \cite{afifi2019image}. Unlike color transfer, color matching emphasizes accurate color reproduction between target and source images depicting the same scene \cite{AN2}.

Color matching is beneficial in various applications within camera pipeline manufacturing and photo editing, including mapping colors between different color spaces (\eg, raw-to-sRGB \cite{ignatov2020replacing} or sRGB-to-raw \cite{punnappurath2021spatially, nam2022learning}), post-capture white-balance editing (\eg, \cite{afifi2019color, afifi2020deep}), reusing camera ISP color modules for previously unsupported cameras through raw-to-raw mapping (\eg, \cite{afifi2021semi, perevozchikov2024rawformer}), ensuring consistent color reproduction across dual-camera systems to improve image quality (\eg, \cite{wu2023efficient, lai2022face, alzayer2023dc2}), and enable seamless dual-camera zoom \cite{wu2025dual}. While access to both source and target images is feasible in certain scenarios (\eg, on-board camera sRGB-to-raw mapping \cite{punnappurath2021spatially} or dual-camera aligned image color mapping \cite{lai2022face, wu2025dual}), there are other situations where this condition is difficult to achieve (\eg, raw-to-raw mapping \cite{afifi2021semi}, raw-to-sRGB mapping \cite{ignatov2020replacing, schwartz2018deepisp}, color mapping to specific embedded photographic styles \cite{AN7, AN8}, or generic color enhancement \cite{liu20234d, conde2024nilut, brateanu2025enhancing}). 

Most existing color matching and transfer techniques, however, assume access to both source and target images or colors (\eg, \cite{AN1, AN2, finlayson2017color, nikonorov2016correcting, fairchild2008matching, reinhard2001color, pitie2005n, xiao2006color, pitie2007automated, rabin2014adaptive, chang2015palette, afifi2021histogan, ding2024regional, ke2023neural}), which limits the applicability of these algorithms.
In this work, we propose cmKAN, a framework suitable for both online paired-based optimization (where both source and target images are available at inference time) and offline training (where the target image is not accessible at inference time). Additionally, our cmKAN can be trained in an unsupervised manner, making it practical for scenarios where obtaining paired training data is challenging. 

Our cmKAN model employs Kolmogorov-Arnold Networks (KANs)~\cite{AN6}, based on the Kolmogorov-Arnold Representation Theorem \cite{kolmogorov:superposition}, for inherent non-linear, spline-based processing of input features, making them particularly well-suited for non-linear color matching~\cite{menesatti2012rgb, tocci2022advantages, suominen2024camera}. KANs use affine combinations of piecewise polynomial, spline-based activation functions, that allow more nuanced and flexible approximations of complex functions. In this work, we show that the KAN layer naturally extends the modern color-matching transform between two pipelines~\cite{AN2}. This motivated us to integrate KANs into our color-matching framework, where we introduce a hypernetwork-based architecture that includes a generator to control the coefficients of the KAN splines. 
Acting as a hypernetwork \cite{ha2016hypernetworks}, the generator produces spatial weights that dictate the behavior of the KAN, enabling dynamic adaptation for color-matching tasks (see Fig.~\ref{fig:teaser}). 

We evaluated our framework on different tasks relevant to camera ISP development and post-capture editing. Specifically, we tested our approach on unsupervised raw-to-raw mapping, where the goal is to transform the raw colors from a source camera to resemble those produced by another camera in its raw space \cite{afifi2021semi}. Additionally, we examined raw-to-sRGB mapping, aiming to map images from the camera raw space to the standard display sRGB space (acting as a learnable camera ISP \cite{zhang2021learning, xing2021invertible, ignatov2020aim} from color perspective) using supervised training. Furthermore, we tested our framework on color mapping in the sRGB space, where the goal is to map the sRGB colors from the source camera to appear consistent with those from the target camera. For this task, we evaluated both unsupervised training (when no paired dataset is available) and supervised training. Additionally, we considered the scenario where both source and target images are available at inference. This situation applies to dual-camera-based processing (\eg, \cite{wu2023efficient, lai2022face, wu2025dual}) or for creating small learned metadata (\ie, in our case, the model's weights) to store for post-capture editing (\eg, \cite{afifi2019color, le2023gamutmlp}).

To evaluate our method's effectiveness in sRGB-to-sRGB color mapping between two different camera ISPs, we introduce a large dataset of 2.5K well-aligned paired images captured by two distinct cameras. We believe this dataset will be a valuable resource for evaluating color-matching techniques in future research. The code, trained models, and dataset will be made publicly available upon acceptance.

\subsection*{Contribution}
\vspace{-2mm}
Our contributions can be summarized as follows:
\begin{itemize}
   \item  We introduce \emph{cmKAN}, a novel lightweight hypernetwork-based 
   color matching framework that leverages KANs. To the best of our knowledge, this is the first work to leverage the spline non-linearity in KANs for highly accurate, and adaptive color matching.

   \item We propose a lightweight generator that predicts spatially varying KAN parameters for localized and content-aware color transformations, featuring novel Illumination Estimator, Color Transformer, and Color Feature Modulator modules. 

   \item We present a large-scale dataset of paired, well-aligned images taken by two different cameras, facilitating the training and evaluation of color-matching methods.

   \item Our experiments show that our method outperforms existing approaches by an average of 37.3\% across multiple tasks (\ie, \emph{raw-to-raw}, \emph{raw-to-sRGB}, and \emph{sRGB-to-sRGB}) for supervised and unsupervised schemes, while remaining lightweight and suitable for on-device use. Additionally, we conducted a user study, demonstrating 2$\times$ superior MOS scores over other methods.

\end{itemize}
\section{Method}
\label{sec:method}
\begin{figure*}[t]
  \centering
  \includegraphics[width=0.91\linewidth]{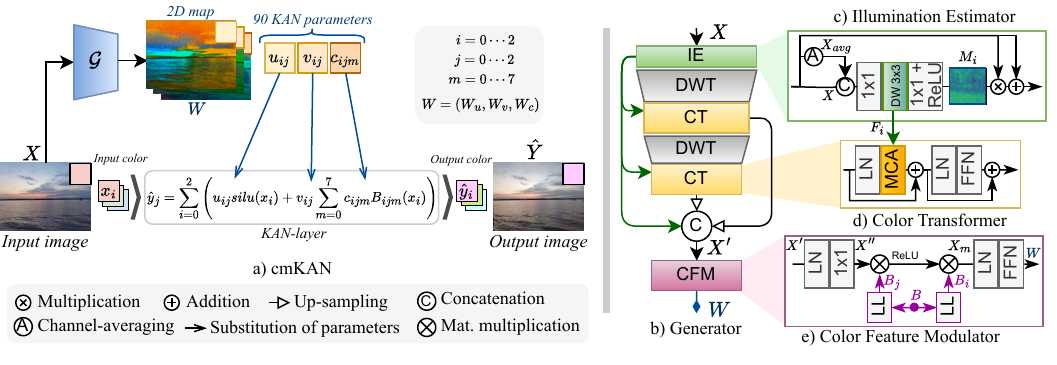}
  \vspace{-4mm}
  \caption{a) \textbf{Overview of the cmKAN architecture}, b) Generator Network ($\mathcal{G}$) and its key components: c) Illumination Estimator (IE), d) Color Transformer (CT), and e) Color Feature Modulator (CFM).}
  \label{fig:overview}
\end{figure*}

\vspace{-1mm}
This section outlines the core principles of our proposed color-matching model using KANs -- cmKAN. As shown in Fig.~\ref{fig:overview} (a), cmKAN primarily comprises a KAN network and a generator. The KAN network adjusts input image colors to match target colors, and the generator (hypernetwork), takes a source input image and produces spatially varying parameter maps for the KAN.

Without loss of generality, we assume the input and target colors come from source and target cameras, respectively, each potentially with its own ISP, causing disparities in the sRGB images we aim to match. The model applies to other scenarios (raw-to-sRGB/raw-to-raw), learning non-linear mapping between color spaces within a camera (raw-to-sRGB) or across cameras (raw-to-raw).  To address these complex mappings, we first detail the core components: KAN (Secs.\ref{sec:kan}) and generator (\ref{sec:generator}).  Then, we discuss offline training (Sec.~\ref{sec:supervised}, Sec.~\ref{sec:unsupervised}) and paired optimization (Sec.~\ref{sec:optimization}) scenario.

\subsection{Color Matching Using KAN}
\label{sec:kan}
\vspace{-1mm}

Color matching maps a source image's color distribution to that of the target image using spatially varying linear, $\mathbf{L}$, and non-linear, $F(\cdot)$, transformations~\cite{AN1, AN3, AN9, AN4}:
\begin{equation}
\label{eq:standart}
    \hat{\textbf{y}} = F(\textbf{x}) \mathbf{L},
\end{equation}
\noindent where $F$ is applied element-wise, and $\textbf{x}$ and $\hat{\textbf{y}}$ are row vectors of input and output colors, respectively. This standard model can be extended using a more complex one with separate functions, $F_{ij}(\cdot)$, for each channel (\textit{see Supp.}):

\begin{equation}
\label{eq:complicated}
\hat{\textbf{y}} = F(\textbf{x}) \mathbf{L} \to \hat{y}_j = \sum\limits_{i=0}^2 F_{ij}(x_i) \cdot l_{ij},
\end{equation}
\noindent where ${x_i}$, ${\hat{y}_j}$ are input and output color components, $l_{ij}$ are elements of $\mathbf{L}$; $i,j=0..2$.

Popular color-matching approaches~\cite{AN1, AN2} using polynomial approximations of the \textit{standard} model (Eq.~\ref{eq:standart}) and even deep CNNs/MLPs often struggle to accurately capture color transformations in intricate cases~\cite{AN4, AN1, afifi2019color}. In contrast, the proposed KAN-based approach leverages trainable splines to describe our \textit{extended} model (Eq.~\ref{eq:complicated}) with greater precision and to provide additional smoothness~\cite{AN6, zeng2024kan} for color correction:

\begin{equation}\label{eq:KAN}
\hat{y}_j = \sum_{i=0}^{2} \left(u_{ij} silu(x_i) + \underbrace{v_{ij}}_{l_{ij}} \underbrace{\sum_{m=0}^{7} c_{ijm} B_{ijm}(x_i)}_{F_{ij}(x_i)}\right)
\end{equation}
\noindent where $x_i$ and $\hat{y}_j$ are the input and output RGB components.  ${silu(.)}$ is a residual activation, $B_{ijm}(.)$ are cubic B-spline basis functions, and $u_{ij}$, $v_{ij}$, $c_{ijm}$ are 90 KAN parameters; $i,j=0..2$, $m=0..7$.  \textit{See Supp. for math. background.}

However, a standard KAN layer (Eq.~\ref{eq:KAN}) operates globally, which restricts its ability to handle localized color mismatches caused by spatially nonuniform illumination and ISP-induced artifacts. To overcome this, we introduce a hypernetwork-driven KAN framework, where a lightweight generator $\mathcal{G}$ dynamically predicts 2D parameter map coding spatially varying KAN parameters. The parameter map is channel-wise split into three components: $W=(W_u,W_v,W_c)$. Each point in $W$ represents 90 non-trainable KAN parameters: $u_{ij}=W_u(\cdot)_{ij}$, $v_{ij}=W_v(\cdot)_{ij}$, $c_{ijm}=W_c(\cdot)_{ijm}$ (Fig.~\ref{fig:overview} (a)). This structured parameterization enables region-specific color transformations while mitigating the adverse effects of noise, ensuring robust and smooth approximations of complex mappings~\cite{zeng2024kan}. Moreover, our method effectively handles high-dynamic-range scenes, mitigating overexposure and underexposure by leveraging distinct generator-driven spline representations tailored to varying illumination conditions. The full color-matching function is then defined as:

\begin{equation}\label{eq:KAN-full}
\hat{Y} = KAN(\mathcal{G}(X,\theta),X),
\end{equation}
\noindent where $KAN$ is the KAN layer (Eq. \ref{eq:KAN}), $\mathcal{G}$ is the generator with trainable parameters $\theta$, and $X$, $\hat{Y}$ are the input and output images.

\subsection{Generator Network}
\label{sec:generator}
\vspace{-1mm}

As shown in Fig.~\ref{fig:overview} (b), the generator architecture efficiently generates 2D parameter maps for the KAN layer, handling color correction, and capturing spatial information for processing non-uniform scene structures. This mirrors the ISP's selective application of color corrections and tone mappings to different image regions. The functionality and contributions of its three specialized modules are briefly described  below.

\begin{figure}
  \centering
  \includegraphics[width=0.8\linewidth]{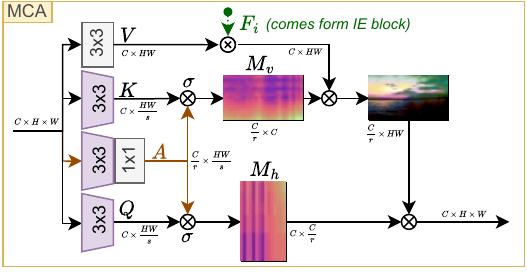}
  \vspace{-2mm}
  \caption{The proposed \textbf{Multi-Scale Color Attention (MCA)} to efficiently exploit the channel-wise dependencies.}
  \label{fig:mca}
\end{figure}

\subsubsection{Illumination Estimator}
\label{sec:IE}
\vspace{-1mm}
The illumination estimator (Fig.~\ref{fig:overview} (c)) provides a foundational step for the overall color processing of the input image $X$. Correct illumination prevents over-saturation or washed-out colors during mapping, preserving the scene's appearance.  Accurate illumination estimation enables precise brightness adjustments, ensuring consistent and realistic color reproduction.

The illumination estimator is a small CNN that processes input image $X$ and its channel-averaged counterpart $X_{avg}$ to generate illumination feature $F_i$ and map $M_i$. It starts with a $1\times1$ convolution to the concatenated $X$ and $X_{avg}$, integrating channel information. A $3\times3$ dilated depth-wise convolution (dilation factor of 2) then expands the receptive field, incorporating contextual information, especially for over/under-exposed regions.  Finally, a $1\times1$ convolution constructs illumination map $M_i$ from feature representation $F_i$. This illumination and features maps guide the multi-scale color attention (MCA) in the subsequent color transformer, refining hidden representations and ensuring improved illumination equalization, preventing over/under-exposed areas.

\subsubsection{Color Transformer}
\label{sec:CT}
\vspace{-1mm}
The Color Transformer (CT) works closely with the Illumination Estimator (IE) for color balance and enhanced spatial coherence. As shown in Fig.~\ref{fig:overview} (d), the CT block uses a ViT-inspired architecture and processes a down-sampled input from the Discrete Wavelet Transform (DWT), preserving boundary details and low-frequency features.  The normalized, down-sampled input and illumination features are processed by the proposed Multi-Scale Color Attention (MCA) for enhanced refinement.  The MCA's normalized output then directly goes to the FFN layer \cite{mst++}, which embeds rich contextual information to augment the features.

\noindent\textbf{Multi-Scale Color Attention:}
The proposed Multi-Scale Color Attention (MCA) (Fig.~\ref{fig:mca}) builds on RawFormer \cite{perevozchikov2024rawformer} with essential enhancements for improved efficiency.  While RawFormer operates spatially, computing similarity between a spatially compressed query ($\mathbf{Q}$) and uncompressed key ($\mathbf{K}$), MCA operates at the channel level. It applies spatial compression to both vectors using strided convolutions (stride $s=2$) and introduces anchors ($\mathbf{A}$) as intermediaries, compressed spatially and channel-wise via strided depthwise separable convolutions. These anchors enable efficient similarity comparison, greatly reducing computational cost without sacrificing accuracy (\textit{See Supplementary}).
Additionally, the $\mathbf{V}$ value vector is modulated by illumination feature $F_i$ to better account for lighting sources, and shifting attention from spatial to channel dimensions helps compute cross-channel covariance, encoding color correction details more efficiently. These changes allow MCA to retain modeling capacity for color reconstruction while significantly reducing complexity.  The MCA operation is summarized as:

\begin{equation}
\begin{array}{c}
    MCA(\textbf{M}_h, \textbf{M}_v, \textbf{V}) = \textbf{M}_h\cdot(\textbf{M}_v\cdot F_i\textbf{V})\\
    \textbf{M}_h = Softmax(\frac{\textbf{Q}\textbf{A}^T}{t_h}), \textbf{M}_v = Softmax(\frac{\textbf{A}\textbf{K}^T}{t_v})\\
\end{array}
\label{eq:mca}
\end{equation}
\noindent where $\textbf{M}_h$ and $\textbf{M}_v$ are horizontal and vertical attention maps from anchor-query and anchor-key pairs, respectively; $t_h$, $t_v$ are their temperatures; and $F_i$ is illumination feature provided by the estimator. To avoid dimension mismatch when multiplying $\textbf{V}$ by $F_i$, $F_i$ is passed through a strided convolution.

\subsubsection{Color Feature Modulator}
\label{sec:CFM}
\vspace{-1mm}
As shown in Fig.~\ref{fig:overview} (b and e), the Color Feature Modulator (CFM) processes concatenated 2D maps ($X'$) from various parts of the generator, a composite multi-scale, color-dependent vector that requires further modulation to generate parameter maps, \( W \), for the KAN, using a mechanism similar to memory colors (common in color correction)~\cite{nikonorov2016correcting}. Thus, to modulate $X'$, we apply linear projections (LL) of trainable bias vectors, $B$, to refine the normalized (LN) concatenated maps, $X''$.  The modulation is: $X_m=B_i\cdot ReLU(X''\cdot B_j)$, where $X''$ is the normalized input, $B_i$, $B_j$ are linear projections (LL) of trainable bias $B$, and $X_m$ is the modulated output. Finally, parameter maps \( W \) are obtained by applying an FFN \cite{mst++} as an output projection to refine $X_m$.

Next, we discuss how the model is adapted for supervised, unsupervised, and paired-based optimization scenarios.

\subsection{Supervised Learning}
\label{sec:supervised}
\vspace{-1mm}
Let \( X \) and \( Y \) represent an input image and its corresponding target image, respectively. For the supervised learning approach, the proposed  cmKAN model generates an estimated image \( \hat{Y} \) from \( X \), aiming to approximate the color characteristics of the target image \( Y \) (Eq.~\ref{eq:KAN-full}). 
The overall framework is trained using the following loss function:
\begin{equation}\label{eq:pix-wise}
\mathcal{L}_{pixel-wise}(Y,\hat{Y}) = \ell_{1} + \beta_0 (1-SSIM)
\end{equation}
where $\beta_0=0.15$, and $\ell_{1}$ and $SSIM$ refer to L1 loss and the structural similarity index measure (SSIM) \cite{wang2004image}, respectively. 

\subsection{Unsupervised Learning}
\label{sec:unsupervised}
\vspace{-1mm}
In our unsupervised learning approach, we adopted two-stage training strategy: (1) generator pre-training and (2) unpaired image-to-image training using a CycleGAN-like framework~\cite{zhu2017unpaired}. For network's pretraining, the input images were divided into $32 \times 32$ patches, and a color transformation including random jitter in the range (-30\%, +30\%) of brightness, contrast, saturation, hue, color shift, and random channel reordering were randomly applied to each patch with 60\% coverage of the image. The overall objective of the generator in  pre-training stage was to reconstruct the original image colors from these partially altered patches via minimizing the pixel-wise loss function in Eq.~\ref{eq:pix-wise}.


Following pre-training, unpaired color matching was performed to jointly train our cmKAN and a discriminator network. Here, we use the discriminator network proposed in \cite{perevozchikov2024rawformer}. 
During this stage, the discriminator networks were optimized to minimize the following loss functions:

\begin{equation}
\label{eq:disc1}
    \mathcal{L}^{dis}_A = \ell_{gan}(D_A(G_{B \rightarrow A}(b)), 0) + \ell_{gan}(D_A(a), 1)
\end{equation}
\begin{equation}
\label{eq:disc2}
    \mathcal{L}^{dis}_B =  \ell_{gan}(D_B(G_{A \rightarrow B}(a)), 0) +  \ell_{gan}(D_B(b), 1)
\end{equation}

\noindent where $a$ and $b$ represent images from the source and target color distributions $A$ and $B$, respectively, while $G_{B \rightarrow A}(b)$ and $G_{A \rightarrow B}(a)$ are the output images produced by each generator network, respectively. The labels $0$ and $1$ represent ``fake'' and ``real'' images, respectively, and $\ell_{gan}(\cdot)$ computes the cross-entropy loss. 

The weights of two cmKAN models -- one mapping from source to target and the other from target to source color distributions -- were optimized to minimize the following loss function:

{\scriptsize\begin{equation}
    \mathcal{L}^{gen} = \beta_1(\mathcal{L}^{gan}_A + \mathcal{L}^{gan}_B) + \beta_2(\mathcal{L}^{idt}_A + \mathcal{L}^{idt}_B) + \beta_3(\mathcal{L}^{cyc}_A + \mathcal{L}^{cyc}_B)
    \label{eq:gen1}
\end{equation}}

\noindent where $\beta_1$, $\beta_2$, and $\beta_3$ where set to 1, 10, 0.5, respectively; $\mathcal{L}^{gan}_A$ and $\mathcal{L}^{idt}_A$  refer to $\ell_{gan}(D_B(G_{A \rightarrow B}(a)), 1)$ and $\mathcal{L}_{pixel-wise}(G_{B \rightarrow A}(a), a)$, respectively. $\mathcal{L}^{cyc}_A$ is $\mathcal{L}_{pixel-wise}( G_{B \rightarrow A}(G_{A \rightarrow B}(a)), a)$. Note that at inference time, only a single cmKAN (e.g., $G_{A \rightarrow B}$) is required to map colors from the source color distribution (e.g., $A$) to the target color distribution (e.g., $B$).

\subsection{Paired-Based Optimization}
\label{sec:optimization}
\vspace{-1mm}

For paired-based optimization, we used pairs of corresponding colors for the source and target images  to align their color distributions effectively. For training between image pairs, we employed cmKAN-Light, a lightweight variant of our full model, optimized for faster performance. 
Unlike the original architecture depicted in Fig.~\ref{fig:overview}, cmKAN-Light is simplified to include only a single Color Transformer (CT) block and one Discrete Wavelet Transform (DWT) block, reducing computational complexity while maintaining essential functionality. 
Given the need for efficient online optimization, we adopted a two-stage approach. First, to provide a robust initialization, we pre-trained the cmKAN-Light model on the corresponding dataset using the supervised learning approach outlined in (Sec.~\ref{sec:supervised}). This step generated a set of well-initialized parameters, serving as a starting point for fine-tuning. Subsequently, fine-tuning was performed directly on the particular image pairs, allowing the model to adapt to specific pairwise relationships.
To further accelerate training, we employed an L1 loss function over just 10 iterations, ensuring rapid convergence. This strategy enables cmKAN-Light to achieve effective color matching with minimal computational overhead, making it well-suited for online optimization tasks.

\section{Experiments}
\label{sec:experiments}
\vspace{-1mm}

In this section, we evaluate our cmKAN across a range of color-matching tasks in different scenarios, including supervised, unsupervised, and paired-based optimization. First, we evaluate our method on unsupervised raw-to-raw mapping, where the goal is to map raw images from a source camera to the raw color space of a target camera. Next, we apply our approach to supervised raw-to-sRGB mapping, where our network learns the ISP non-linear color mapping function. Finally, we evaluate cmKAN on sRGB-to-sRGB mapping, to transfer colors from a source color distribution to a target distribution, such as professional photographer edits \cite{AN7} or another camera ISP rendering \cite{AN2}. Additionally, we test our method on our proposed dataset of paired sRGB images rendered by the camera ISP and captured by two different cameras. For this sRGB-to-sRGB mapping task, we present results across the three scenarios: unsupervised, supervised, and paired-based optimization.

\subsection{Datasets}
\vspace{-1mm}

We begin by introducing our dataset, collected to support the training and evaluation of sRGB-to-sRGB mapping for images captured by two different cameras. Next, we provide an overview of the datasets used throughout our evaluation, organized by each specific task.
\subsubsection{Proposed dataset}
\label{sec:proposed-datset}
\vspace{-1mm}

To validate our cross-camera mapping in sRGB color space, we collected a dataset of 2,520 sRGB images captured by the wide and main cameras (1,260 images per camera) of the Huawei P40 Pro smartphone. These cameras use distinct sensor types: a Quad-Bayer RGGB (Sony IMX700) for the wide camera and an RYYB (Sony IMX608) for the main camera. We used the on-device camera ISP to render images to the sRGB space. sRGB image pairs were then spatially aligned by first identifying keypoints with the SURF detector \cite{bay2006surf} and matching features using the KNN algorithm with Lowe's ratio test \cite{lowe2004distinctive}. A projective transformation matrix was subsequently applied to warp the images. Since both cameras are fixed within the smartphone, this stable positioning helped to significantly minimize misalignment errors. Finally, all aligned images were manually reviewed to discard any remaining misaligned pairs. Our dataset reflects real-world conditions, with images captured under diverse lighting and seasonal variations \textit{(see supp. materials for additional details).}

\subsubsection{Raw-to-raw mapping}
\vspace{-1mm}

For this task, we utilized the raw-to-raw mapping dataset \cite{afifi2021semi}, which contains a total of 392 unpaired raw images (196 from each of the Samsung Galaxy S9 and iPhone X). Additionally, the dataset includes 115 paired testing raw images from each camera for evaluation.

\subsubsection{Raw-to-sRGB mapping}
\vspace{-1mm}

For this task, we utilized the Zurich raw-to-sRGB dataset~\cite{ignatov2020replacing} to evaluate our method on the raw-to-sRGB task. This dataset consists of 48,043 paired raw images and their corresponding sRGB-rendered outputs, captured simultaneously with a Huawei P20 smartphone camera and a Canon 5D Mark IV DSLR camera. For our experiments, we used 70\% of the images for training and 30\% for testing.

\subsubsection{sRGB-to-sRGB mapping}
\vspace{-1mm}

As mentioned previously, in the sRGB-to-sRGB task, we apply color matching to emulate specific photographer styles, replicate the color rendering of target camera ISPs, and align colors between sRGB images captured by two different cameras within a dual-camera device. For the photographer-style matching, we used the MIT-Adobe FiveK dataset \cite{AN7}, which provides expert-adjusted ground-truth styles. We specifically chose expert style C as our target, as it is widely referenced in prior work \cite{chen2018deep, wang2019underexposed, afifi2021learning}. In our experiments, we employed 5K images for supervised training and 498 images for evaluation. 

For camera rendering mapping, we conducted paired-based optimization evaluations using the dataset from \cite{AN2}, which includes 35 images captured by two DSLR cameras (Nikon D3100 and Canon EOS 80D) from the same scenes with slightly different camera positions. This dataset provides synthetic ground-truth images to support color rendering transfer from the source to the target camera, ensuring well-aligned ground-truth pairs.

Finally, we used our proposed  dual-camera sRGB dataset to evaluate all the three scenarios: supervised, unsupervised, and paired-based. For both the supervised and unsupervised experiments, 900 images were allocated for training and 350 images for testing.




\subsection{Training details}
\vspace{-1mm}

We used the following patch sizes for training: 256$\times$256 patches for raw-to-raw mapping, 448$\times$448 patches for raw-to-sRGB, and 1024$\times$1024 patches for sRGB-to-sRGB mapping. In all the experiments, we used Adam optimizer~\cite{kingma2014adam} for supervised training, unsupervised training, and paired-based optimization with betas set to (0.9, 0.99).

For unsupervised training, we first trained our cmKAN on the color reconstruction task (as explained in Sec.~\ref{sec:unsupervised}) for 200 epochs, and then trained it on the unpaired data for the color-matching task for an additional 500 epochs. We used a learning rate of 0.0001 for the discriminator networks to minimize Eqs.~\ref{eq:disc1} and \ref{eq:disc2}, and a learning rate of 0.001 for the cmKAN models to minimize Eq.~\ref{eq:gen1}.  For supervised training (i.e., raw-to-sRGB and supervised sRGB-to-sRGB), we trained our model for 450 epochs to minimize the loss function in Eq.~\ref{eq:pix-wise}. For both supervised training and paired-based optimization, we used a learning rate of 0.001.

\subsection{Results}
\vspace{-1mm}

\subsubsection{Raw-to-raw mapping results}
\vspace{-1mm}

\begin{figure}[t]
  \centering
  \includegraphics[width=0.8\linewidth]{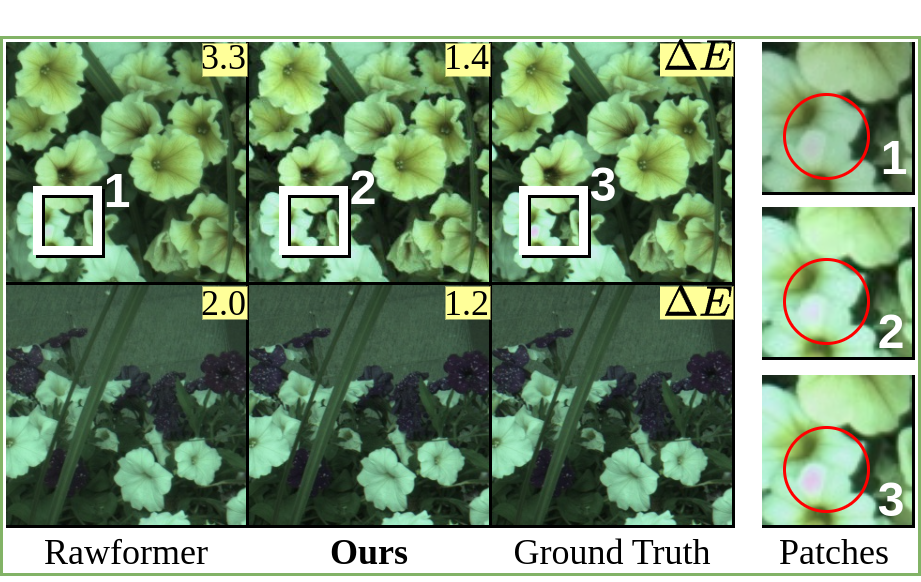}
  \vspace{-2mm}
  \caption{
  Qualitative comparison of unsupervised \textit{raw-to-raw mapping} on the
dataset in \cite{afifi2021semi}. Our cmKAN method achieves the most accurate color mapping with lower $\Delta E$ errors, while being more lightweight than RawFormer~\cite{perevozchikov2024rawformer}. Gamma operator is applied to aid visualization. Best viewed in the electronic version.}
  \vspace{-2mm}
  \label{fig:qualtative-raw2raw}
\end{figure}

We compare our unsupervised-trained model for raw-to-raw mapping against the following methods (trained in an unsupervised or semi-supervised manner as described in \cite{perevozchikov2024rawformer} and \cite{afifi2021semi}, respectively): SSRM \cite{afifi2021semi}, UVCGANv2 \cite{torbunov2023uvcgan2}, and RawFormer \cite{perevozchikov2024rawformer}. We report PSNR, SSIM, and $\Delta E$ errors in Table \ref{tab:raw-to-raw} for the raw-to-raw mapping dataset \cite{afifi2021semi}. As shown, our method achieves better or comparable results to the recent work in \cite{perevozchikov2024rawformer}, while being lightweight and more compatible with limited computational resources. Qualitative examples demonstrating the efficacy of the proposed method are shown in  Fig.~\ref{fig:qualtative-raw2raw} \textit{(more in supp. materials).}

\begin{table}[h]
  \centering
    \caption{Results for \textit{unsupervised} raw-to-raw mapping using the dataset in \cite{afifi2021semi}. The best results are highlighted in yellow. \label{tab:raw-to-raw}}
  \resizebox{0.8\linewidth}{!}
  {
  \begin{tabular}{@{}l|ccc|ccc@{}}
    \toprule
    Method & PSNR & SSIM & $\Delta E$ & PSNR & SSIM & $\Delta E$\\
    & \multicolumn{3}{c|}{\colorbox{supervised}{Samsung-to-iPhone}} & \multicolumn{3}{c}{\colorbox{unsipervised}{iPhone-to-Samsung}}\\
    \midrule
    SSRM~\cite{afifi2021semi}    & 29.65 & 0.89 & 6.32 & 28.58 & 0.90 & 6.53 \\
UVCGANv2~\cite{torbunov2023uvcgan2}  & 36.32 & 0.94 & 4.21 & 36.46 & 0.92 & 4.73 \\
    RawFormer~\cite{perevozchikov2024rawformer}& 40.98 & \colorbox{best}{0.97} & 2.09 & \colorbox{best}{41.48} & \colorbox{best}{0.98} & 1.99 \\
    cmKAN     & \colorbox{best}{41.01} & \colorbox{best}{0.97} & \colorbox{best}{1.23} & 41.47 & \colorbox{best}{0.98} & \colorbox{best}{1.27} \\
    \bottomrule
  \end{tabular}
  }  \vspace{-2mm}
  \label{tab:raw}
\end{table}

\subsubsection{Raw-to-sRGB mapping results}
\vspace{-1mm}

\begin{figure}[t]
  \centering
  \includegraphics[width=0.9\linewidth]{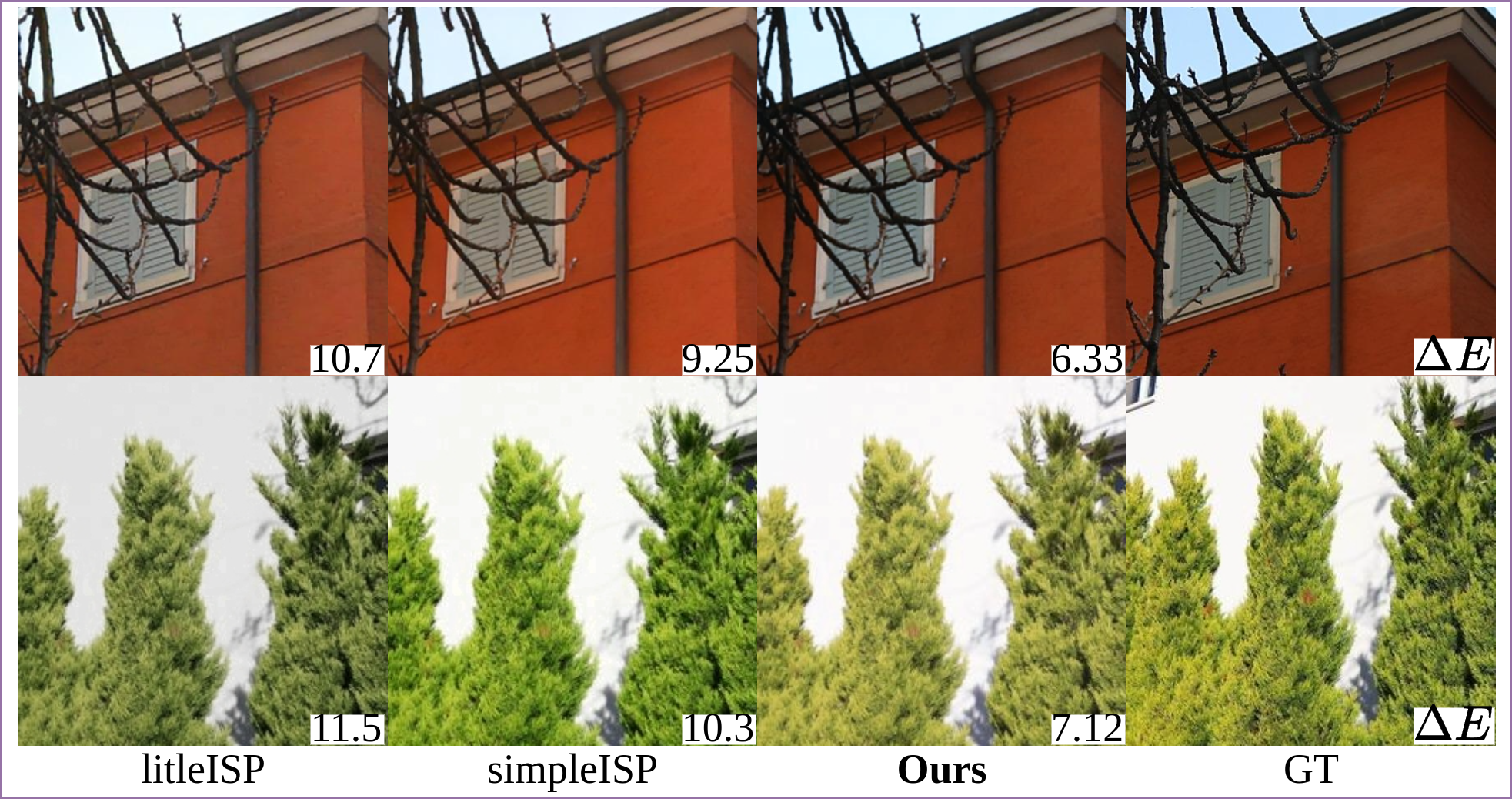}
    \vspace{-2mm}
  \caption{Qualitative comparison of raw-to-sRGB rendering on the Zurich raw-to-sRGB dataset~\cite{ignatov2020replacing}. Our cmKAN method achieves the most accurate color mapping with lower $\Delta E$ errors. Best viewed in the electronic version.}
  \vspace{-2mm}
  \label{fig:qualtative-isp}
\end{figure}

We compare our method for learning raw-to-sRGB color rendering in camera ISPs against the following methods (all trained in a supervised manner on the same dataset): 
MW-ISP~\cite{ignatov2020aim}, LiteISP~\cite{zhang2021learning}, MicroISP~\cite{ignatov2022microisp} , and SimpleISP~\cite{elezabi2024simple}. Table \ref{tab:isp} reports PSNR, SSIM, and $\Delta E$ error 
metrics demonstrating that our method achieves promising results with reduced computational cost (FLOPs). Qualitative examples are shown in Fig.~\ref{fig:qualtative-isp} \textit{(more in supp. materials)}.

\begin{table}[t]
  \centering
    \caption{Results for \textit{supervised} raw-to-sRGB mapping on the Zurich raw-to-sRGB dataset~\cite{ignatov2020replacing}. }
      \label{tab:isp}
          \vspace{-2mm}
  \resizebox{0.65\linewidth}{!}
  {
  \begin{tabular}{@{}l|ccc|c@{}}
    \toprule
    Method & PSNR & SSIM & $\Delta E$ & FLOPs \\
    \midrule
    MW-ISP~\cite{ignatov2020aim}  & 21.88 &  0.82 & 10.33 &  3.6T\\
    LiteISP~\cite{zhang2021learning} & 22.18 &  0.83 & 10.28 & 174G\\
    MicroISP~\cite{ignatov2022microisp} & 20.30 & 0.78 & 11.14 & \colorbox{best}{37G}\\
    SimpleISP~\cite{elezabi2024simple} & 24.18 & 0.84 & 10.02 & 58G\\
    cmKAN   & \colorbox{best}{24.41} & \colorbox{best}{0.85} &  \colorbox{best}{7.27} & 40G\\
    \bottomrule
  \end{tabular}
  \vspace{-5mm}
  }
\end{table}

\subsubsection{sRGB-to-sRGB mapping results}
\vspace{-1mm}

\begin{figure*}[t]
  \centering
  \includegraphics[width=0.9\linewidth]{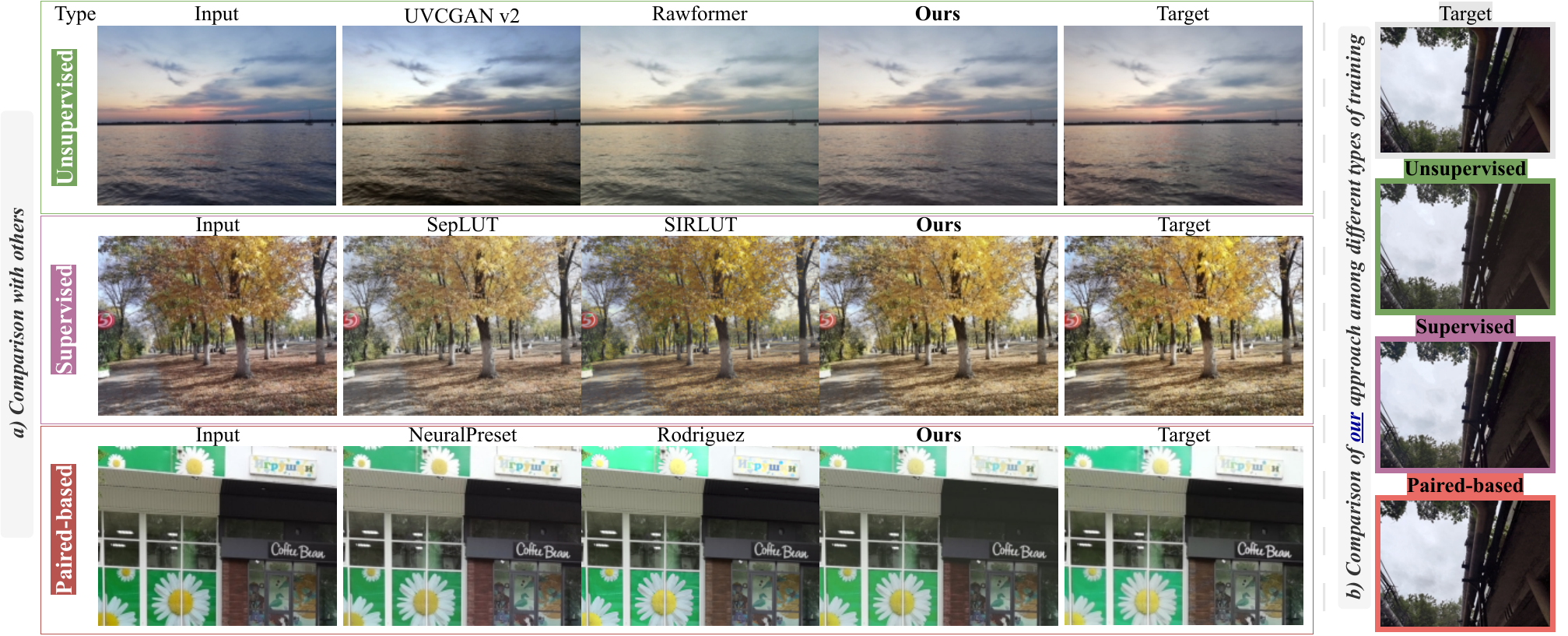}
  \vspace{-1mm}
  \caption{Qualitative comparison of sRGB-to-sRGB color matching on our dataset. In (a), we show three scenarios using our method: unsupervised learning (green), supervised learning (purple), and paired-based optimization (red). Our method demonstrates superior color matching compared to other methods (UVCGAN v2~\cite{torbunov2023uvcgan2}, Rawformer~\cite{perevozchikov2024rawformer}, SepLUT~\cite{yang2022seplut}, SIRLUT~\cite{li2024sirlut}, NeuralPreset~\cite{ke2023neural}, and Rodriguez \textit{et al}.'s method~\cite{AN2}). 
  In (b), we show results from a single scene across the three scenarios using our method, showcasing consistent color-matching accuracy.}
  \vspace{-3mm}
  \label{fig:qualtative-srgb2srgb}
\end{figure*}

In Table~\ref{tab:cm-ours}, we report the results on our proposed dataset in three different scenarios. Firstly,  we compare our unsupervised-trained model against other unsupervised methods, namely UVCGANv2~\cite{torbunov2023uvcgan2} and RawFormer~\cite{perevozchikov2024rawformer}. Then, we compare our supervised-trained model against several methods trained in a supervised manner, including: 
SepLUT~\cite{yang2022seplut}, 
MW-ISP~\cite{ignatov2020aim},
LYT-Net~\cite{brateanu2024lyt}, and SIRLUT~\cite{li2024sirlut}. Lastly, we compare our paired-based optimization results against other paired-based methods for color matching and transfer. Specifically, we compare our results with the following methods: polynomial mapping~\cite{hong2001study}, root-polynomial mapping~\cite{finlayson2015color}, 
NeuralPreset~\cite{ke2023neural}, and Rodriguez et al.'s method~\cite{AN2}. 
As shown, our method achieves the best results across all metrics, while requiring less inference time and having fewer parameters compared to the other methods. Qualitative comparisons for the same are shown in Fig.~\ref{fig:qualtative-srgb2srgb} \textit{(more results in supp. materials)}.

Lastly, the results on the MIT-Adobe FiveK~\cite{AN7}  dataset for supervised sRGB-to-sRGB mapping to emulate a specific photographer style, are presented in Table \ref{tab:cm-5k} (\textit{results on the PPR10K dataset~\cite{AN8}
for the same task are in supp. materials}). 

The results clearly show that our method consistently outperforms alternative approaches across various color-matching tasks. Additionally, we validated our results by conducting a user study that shows \textbf{\textit{2$\times$ superior MOS scores}} over other methods \textit{(see supp. for more details)}.

\begin{table}[t]
  \centering
    \caption{Results of sRGB-to-sRGB mapping on our dataset. We compare the results of \textit{unsupervised} trained models, \textit{supervised} trained models, and \textit{paired-based inference}. Inference times were measured on GPU (except for methods marked by *, which were run on CPU).}
    \vspace{-1mm}
  \label{tab:cm-ours}
  \resizebox{0.9\linewidth}{!}
  {
  \begin{tabular}{@{}l|ccc|cc@{}}
    \toprule
    Method & PSNR & SSIM & $\Delta E$ & \#Params & Time\\
    \midrule
     & \multicolumn{3}{c|}{\colorbox{unsipervised}{Unsupervised training}} & \\
    UVCGANv2~\cite{torbunov2023uvcgan2}& 22.89 & 0.74 & 6.44 & 32.6M & 2.1s\\
    RawFormer~\cite{perevozchikov2024rawformer}& 23.81 & 0.77 & 5.78 & 26.1M & 6.7s\\
    cmKAN  & \colorbox{best}{25.64} &  \colorbox{best}{0.88} &  \colorbox{best}{4.86} &   \colorbox{best}{76.4K} &  \colorbox{best}{1.1s}\\
    \midrule
     & \multicolumn{3}{c|}{\colorbox{supervised}{Supervised training}} & \\
    SepLUT~\cite{yang2022seplut}        & 22.86 & 0.74 & 6.31 & 119.8K & 2.1s\\
    MW-ISP~\cite{ignatov2020aim}        & 23.31 & 0.76 & 5.93 & 29.2M & 8.8s\\
    LYT-Net~\cite{brateanu2024lyt}       & 24.19 & 0.78 & 5.64 &  1.1M & 3.9s\\
    SIRLUT~\cite{li2024sirlut}        & 24.12 & 0.78 & 5.59 & 113.3K & 2.1s\\
    cmKAN  & \colorbox{best}{25.94} & \colorbox{best}{0.89} & \colorbox{best}{4.51} & \colorbox{best}{76.4K} & \colorbox{best}{1.1s}\\
    \midrule
     & \multicolumn{3}{c|}{\colorbox{paired}{ Paired-based inference}} & \\
    Poly~\cite{hong2001study}          & 24.27 & 0.78 & 5.08 & 57 & \colorbox{best}{0.8s*}\\ 
    Root-poly~\cite{finlayson2015color}    & 24.53 & 0.80 & 4.92 & 120 & 4.6s*\\ 
    NeuralPreset~\cite{ke2023neural}  & 23.35 & 0.76 & 5.97 & 5.15M & 20.4s*\\
    R.G. Rodriguez et al. ~\cite{AN2}& 24.35 & 0.78 & 4.99 & \colorbox{best}{14} & 3.3s*\\
     cmKAN-Light   & \colorbox{best}{24.57} & \colorbox{best}{0.81} & \colorbox{best}{4.79} & 7.8K & 1.5s\\
    \bottomrule
  \end{tabular}  \vspace{-1mm}
  }
\end{table}


\begin{table}[t]
  \centering
    \caption{Results of \textit{supervised} sRGB-to-sRGB mapping on the MIT-Adobe FiveK dataset~\cite{AN7}. The best-performing results are highlighted in yellow.}
        \vspace{-1mm}
  \resizebox{0.55\linewidth}{!}
  {
  \begin{tabular}{@{}l|ccc@{}}
    \toprule
    Method & PSNR & SSIM & $\Delta E$\\
    \midrule
    SepLUT~\cite{yang2022seplut}        & 25.47 & 0.92 & 7.54\\
    LYT-Net~\cite{brateanu2024lyt}       & 24.10 & 0.92 & 7.03\\
    SIRLUT~\cite{li2024sirlut} & 27.25 & 0.94 & 6.19\\
    cmKAN  & \colorbox{best}{31.74} & \colorbox{best}{0.95} & \colorbox{best}{2.83}\\ 
    \bottomrule
  \end{tabular}  
  \vspace{-1mm}
  }
  \label{tab:cm-5k}
\end{table}

\subsection{Ablation Studies}
\vspace{-1mm}

\begin{figure}[t]
  \centering
  \includegraphics[width=1.\linewidth]{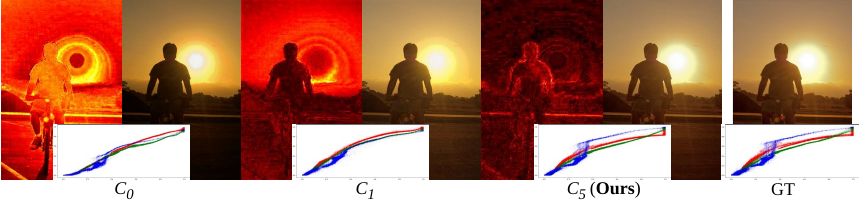}
    \vspace{-1mm}
  \caption{Visualization of results on Adobe FiveK~\cite{AN7} dataset using the crucial configurations, including $\Delta E$ error maps and scatter plots between input and output colors. The synergy of the KAN layer and 2D maps of parameters enables the construction of distinct splines for different regions, enhancing performance in challenging lighting conditions.}
  \vspace{-1mm}
  \label{fig:errors}
\end{figure}

The ablation studies for supervised sRGB-to-sRGB on our dataset (Table \ref{tab:a-studies}) validate our design choices by systematically assessing key components — the KAN and the Generator network. As a baseline ($C_0$), we replace KAN with a three-layer MLP (3,13,3) containing 90 parameters for color transformation. Additionally, the generator uses Rawformer's basic blocks — CQA and SPFN~\cite{perevozchikov2024rawformer} instead of the proposed IE, MCA, and CFM blocks and finally generates a 1D global vector via average pooling instead of 2D spatial parameter maps. Replacing the MLP with a 1D KAN in $C_1$ leads to a 2.29 dB improvement, highlighting the effectiveness of KAN for color transformation (as evident from error maps and scatter plots in Fig.~\ref{fig:errors}). In $C_2$, removing average pooling and allowing the generator to output 2D spatial maps further improves the performance. Incrementally incorporating our proposed components—Illumination Estimator ($C_3$), Color Transformer with MSA ($C_4$), and Color Feature Modulator ($C_5$)—yields additional gains of 2.9 dB, 2.98 dB, and 3.33 dB, respectively. \textit{See supp. for more ablations and architectural details of different configurations}.

\begin{table}[t]
  \centering
  \caption{Ablation study on the impact of different components of our method and various network architectures. Results are reported on our dataset.}
      \vspace{-2mm}
  \resizebox{0.9\linewidth}{!}
  {
  \begin{tabular}{@{}lcccccc|ccc@{}}
    \toprule
    Config & Baseline & KAN & 2D Map & IE & CT & CFM & PSNR & SSIM & $\Delta E$ \\
    \midrule
    $C_0$ & \checkmark &            &            &            &            &            & 22.61 & 0.74 & 6.41 \\
    $C_1$ & \checkmark & \checkmark &            &            &            &            & 24.90 & 0.84 & 5.10 \\
    $C_2$ & \checkmark & \checkmark & \checkmark &            &            &            & 25.23 & 0.86 & 4.71 \\
    $C_3$ & \checkmark & \checkmark & \checkmark & \checkmark &            &            & 25.51 & 0.87 & 4.65 \\
    $C_4$ & \checkmark & \checkmark & \checkmark & \checkmark & \checkmark &            & 25.59 & 0.87 & 4.58 \\
    $C_5$ (\textbf{ours}) & \checkmark & \checkmark & \checkmark & \checkmark & \checkmark & \checkmark &  \colorbox{best}{25.94} &  \colorbox{best}{0.89} & \colorbox{best}{4.51} \\
    \bottomrule
  \end{tabular}
  }  \vspace{-2mm}
  \label{tab:a-studies}
\end{table}

\section{Conclusion}
\label{sec:conclusion}
\vspace{-1mm}

We have presented \emph{cmKAN}, a versatile color-matching framework based on the Kolmogorov-Arnold Network (KAN). Our method includes a KAN layer, controlled by spatially adaptive weights produced by an efficient generator network, which achieves target-matching colors in the final output image. We demonstrate three applications of our framework: supervised training, unsupervised training, and paired-based optimization. We validated our approach across multiple color-matching tasks, including 1) \emph{raw-to-raw mapping:} transferring colors from a source camera's raw space to match a target camera's raw space; 2) \emph{raw-to-sRGB mapping:} aimed at learning the color rendering process of a camera’s ISP, and 3) \emph{sRGB-to-sRGB mapping:} aligning the sRGB colors of source images to match colors from another camera or a photographer's style. Our \emph{cmKAN} achieves SoTA results, outperforming other methods while remaining lightweight and computationally efficient.


\maketitlesupplementary
\textit{
In this supplementary material, we first discuss the mathematical background of applying Kolmogorov-Arnold Networks (KANs) to the color-matching task in Sec.~\ref{sec:splines}. Next, we provide additional ablation studies in Sec.~\ref{sec:extra-ablation}. Then, we elaborate on our proposed dataset in Sec.~\ref{sec:dataset-details}. In Sec.~\ref{sec:extra-results}, we present and results of our method for raw-to-raw, raw-to-sRGB, and sRGB-to-sRGB tasks. Finally, in Sec.~\ref{sec:user-study}, we describe and analyze the user study conducted to validate our method from a human guidance perspective.
}

\section{KANs and Color-Matching Problem}
\label{sec:splines}
This section details the mathematical reasoning for using a single KAN layer in our color-matching task.

An abstract camera Image Signal Processing (ISP) pipeline can be represented as linear and non-linear transformations~\cite{AN1, AN3, AN9, AN4, finlayson2017color}. Specifically, the image formation process can be represented as:  
\begin{equation}\label{eq:isp}
{I}_{\text{rgb}} = T({I}_{\text{raw}} \mathbf{L}),
\end{equation}
\noindent where ${I}_{\text{raw}}$ is the input image in the camera raw space, ${I}_{\text{rgb}}$ is the output in one of the display standard spaces (\textit{e.g.}, sRGB), $\mathbf{L}$ represents the linear component of the ISP pipeline, and ${T}$ implements the non-linear transformations (tone mapping, gamut mapping, image enhancement, etc.)~\cite{zhang2023lookup}. 

For color matching between two images of the same scene processed through different ISPs ($ISP_x$ and $ISP_y$), the color matching of two pixels $\textbf{x}$ and $\textbf{y}$ in the corresponding images, $X = ISP_x(I_\text{raw})$ and $Y = ISP_y(I_\text{raw})$, is:
\begin{equation}\label{eq:translation}
\hat{\textbf{y}} = T_{y}(T_{x}^{-1}(\textbf{x}) \mathbf{L}),
\end{equation}
\noindent where $\mathbf{L}$ represents a linear transformation, and ${T}_{x}$ and ${T}_{y}$ are the non-linear color transformations of the respective ISPs. Note that $T_{x}$ and $T_{y}$ are applied element-wise; $\textbf{x}$ and $\hat{\textbf{y}}$ are row vectors.

Recent research in color matching~\cite{AN1} (Equation 7) shows that the non-linear transformations $T_{x}$ and $T_{y}$ can be represented by a single operation $F$, and the linear part $\mathbf{L}$ can be factored out.  Thus, we can rewrite Equation~\ref{eq:translation} as:
\begin{equation}\label{eq:tranlation}
\hat{\textbf{y}} = F(\textbf{x}) \mathbf{L},
\end{equation}
\noindent where $F$ is applied element-wise; $\textbf{x}$ and $\hat{\textbf{y}}$ are row vectors.

Accurate approximation of $F(\cdot)$ requires a plausible parametric space. Current state-of-the-art methods~\cite{AN1, finlayson2017color} utilize polynomial approximations of Eq.~\ref{eq:tranlation}. Although these methods may outperform deep CNNs/MLPs in color matching~\cite{AN4, AN1}, they can struggle to accurately represent color transformations in intricate cases~\cite{AN4, AN1, afifi2019color}. To address this, we propose using a more complex model that incorporates separate functions $F_{ij}(\cdot)$ for each channel:
\begin{equation}
\label{eq:complicated}
\hat{\textbf{y}} = F(\textbf{x}) \mathbf{L} \to \hat{y}_j = \sum\limits_{i=0}^2 F_{ij}(x_i) \cdot l_{ij},
\end{equation}
\noindent where ${x_i}$, ${\hat{y}_j}$ are input and output color components, $l_{ij}$ are elements of $\mathbf{L}$; $i,j=0..2$.

To enable the use of KANs for color-matching, we propose parameterizing the non-linear components, $F_j(\cdot)$, with a B-spline approximation of Eq.~\ref{eq:complicated}:
\begin{equation}\label{eq:splines_for_color}
\hat{y}_j = \sum_{i=0}^{2} \left(\sum_{m=0}^{G+k-1} c_{ijm} B_{ijm}(x_i)\right) \cdot l_{ij},
\end{equation}
\noindent where ${x_i}$, ${\hat{y}_j}$ are input and output color components, $l_{ij}$ are elements of $\mathbf{L}$, $c_{ijm}$ are spline coefficients and $B_{ijm}(\cdot)$ are B-spline basis functions of order $k$ and grid size $G$; $i,j=0..2$.

This B-spline approximation can be directly implemented with a \textit{single KAN layer}~\cite{AN6}, offering several advantages. KAN is comparable to MLP and easily integrates with neural networks. Unlike MLPs (based on the Universal Approximation Theorem), KANs use the Kolmogorov-Arnold Representation Theorem \cite{kolmogorov:superposition}, with learnable activation functions on edges instead of weights, and the summation of the resultant learned function's output is performed at the nodes.  Additionally, the KAN layer provides smooth spline approximation, crucial for accurate color matching~\cite{zeng2024kan, AN6}.  In our implementation, we use a single KAN layer~\cite{AN6} which uses a residual connection and expresses the activation function as a combination of \(silu(x)\) and a B-spline, with 3 inputs and outputs, spline order $k = 3$, and grid size $G = 5$, represented as:
{\scriptsize
\begin{equation}
\hat{y}_j = \sum_{i=0}^{2} \left(u_{ij} silu(x_i) + \underbrace{v_{ij}}_{l_{ij}} \underbrace{\sum_{m=0}^{G+k-1} c_{ijm} B_{ijm}(x_i)}_{F_{ij}(x_i)}\right)
\label{eq:kan_layer}
\end{equation}
}
\noindent where ${x_i}$, ${\hat{y}_j}$ are input and output components, ${silu(\cdot)}$ is the residual activation, $B_{ijm}(.)$ are B-spline basis functions, $u_{ij}$, $v_{ij}$, $c_{ijm}$ are KAN layer parameters, $i,j=0..2$, and $G$ and $k$ are grid size and spline order.  Note that unlike ~\cite{AN6}, our KAN layer has no trainable parameters ($u_{ij}$, $v_{ij}$, $c_{ijm}$); the hypernetwork (generator network) dynamically provides them for each pixel.

\section{Ablation Studies}
\label{sec:extra-ablation}

\begin{figure}[!th]
  \centering
  \includegraphics[width=1.\linewidth]{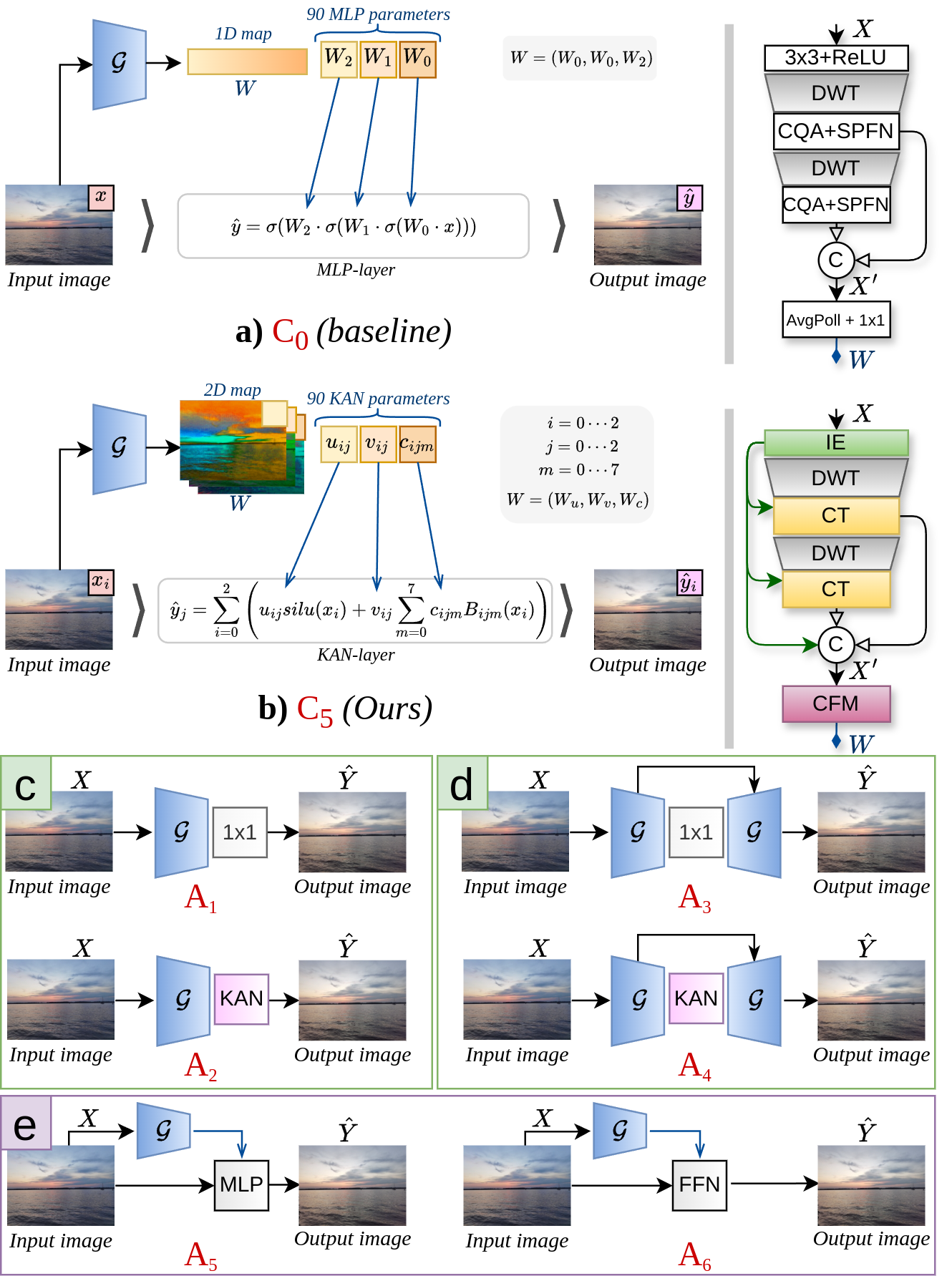}
  \vspace{-3mm}
  \caption{Visualization of various ablation studies architectures: (a) and (b) represent the baseline ($C_0$) and our approach ($C_1$); c) the generator with ($A_1$) and without ($A_0$) fixed learnable KAN layer; d) U-Net-like architecture with ($A_2$) and without ($A_3$) a KAN layer in the bottleneck; e) hyper-networks with MLP ($A_4$) and FFN ($A_5$).}
  \label{fig:a-studies-arch}
\end{figure}

\begin{table}[!ht]
\centering
\caption{\textit{Ablation results} of the impact of the compression operation on vectors of attention on the proposed dataset.}
\label{tab:study_kqv}
\begin{tabular}{c|ccc|c}
\toprule
Modification & PSNR& SSIM & $\Delta E$ & GFLOPs   \\
\midrule
w/o compression & 25.92  & 0.89 & 4.57 & 7.99 \\
w compression   & \colorbox{best}{25.94}  & \colorbox{best}{0.89} & \colorbox{best}{4.51} & \colorbox{best}{2.94} \\

\bottomrule
\end{tabular}
\end{table}

In the main paper, we presented several ablation studies conducted to validate the decisions made in the proposed design of cmKAN (Fig.~\ref{fig:a-studies-arch}). Here, we present an additional ablation experiment performed to validate the operations of the Multi-Scale Color Attention (MCA) block and general architecture approach. 

\begin{table}[!ht]
  \centering
    \caption{Ablation study on the impact of various architectures and the KAN layer. Results of sRGB-to-sRGB mapping on our dataset.}
    \vspace{-2mm}
  \label{tab:a-studies-arch}
  \resizebox{1.0\linewidth}{!}
  {
  \begin{tabular}{@{}l|ccc|cc@{}}
    \toprule
    Method & PSNR & SSIM & $\Delta E$ & \#Params\\
    \midrule
     & \multicolumn{3}{c|}{\colorbox{unsipervised}{a) Embedded KAN}} & \\
    $A_0$ Generator     & 23.81 & 0.78 & 5.79 & 71.3K \\
    $A_1$ Generator+KAN & 24.67 & 0.84 & 5.11 & 76.8K \\
    $A_2$ U-Net         & 24.33 & 0.81 & 5.30 & 117.7K \\
    $A_3$ U-Net+KAN     & 24.89 & 0.85 & 5.09 & 107.7K \\
    \midrule
     & \multicolumn{3}{c|}{\colorbox{supervised}{b) Hyper Networks}} & \\
    $A_4$ Generator+MLP & 24.08 & 0.78 & 5.69 & 115.9K \\
    $A_5$ Generator+FFN & 24.22 & 0.79 & 5.41 & 82.1K \\
    \midrule
    \textbf{Ours}  & \textbf{\colorbox{best}{25.94}} & \textbf{\colorbox{best}{0.89}} & \textbf{\colorbox{best}{4.51}} & \textbf{\colorbox{best}{76.4K}}\\
    \bottomrule
  \end{tabular}  \vspace{-6mm}
  }
\end{table}

Table~\ref{tab:study_kqv} demonstrates the impact of spatial-wise compression, on the attention vectors ($Q$, $K$, and $A$) of the MCA block. This ablation study demonstrates that incorporating compression into these vectors improves performance across all quantitative metrics with significantly lower computational complexity (2.94 $vs$ 7.99) due to reduced matrix dimension.

To validate the effectiveness of our KAN hyper-network approach, we compared our method against various architecture types (Table \ref{tab:a-studies-arch}). \textit{Direct RGB Prediction} ($A_0$): We modified the generator to directly predict a 3-channel RGB vector instead of emitting the KAN 90 parameters. \textit{Generator with Fixed Learnable KAN Layer} ($A_1$): We applied a learnable KAN layer after the generator, without parameter substitution. \textit{Base U-Net Architecture} ($A_2$): U-Net with Color Transformer blocks and DWT/IWT for down- and up-sampling in the encoder and decoders. \textit{U-Net with KAN Integration} ($A_3$): We replaced the U-Net bottleneck with KAN, using a 90-dimensional input to predict RGB values. Finally, we evaluate the impact of using an MLP instead of KAN by replacing KAN with two alternative configurations: a large MLP with five layers and dimensions (3, 12, 24, 12, 3) ($A_4$), and an FFN~\cite{mst++} ($A_5$). The results in Table~\ref{tab:a-studies-arch} show that while KAN layers ($A_1$, $A_3$) improve results, our approach yields superior performance, because the generator provides parameters for the KAN layer, ensuring their smoothness and increasing the robustness of the KAN layer application to noisy data~\cite{zeng2024kan, AN6}.

\section{Additional Details of Our Dataset} \label{sec:dataset-details}

\begin{figure*}[!ht]
  \centering
  \includegraphics[width=1.\linewidth]{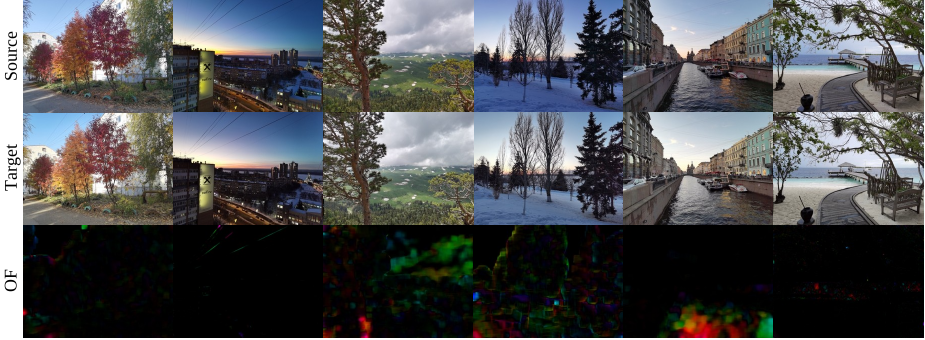}
  \vspace{-3mm}
  \caption{Example images from our dataset alongside their corresponding optical flow (OF) maps. The dataset spans diverse locations and lighting conditions, featuring well-aligned image pairs with minimal matching errors.}
  \label{fig:dataset}
\end{figure*}

In the main paper, we presented our large-scale dataset captured using a Huawei P40 Pro phone.  This device was specifically chosen because it features two distinct cameras with different sensor types: Quad-Bayer RGGB sensor (Sony IMX700) and RYYB sensor (Sony IMX608). These differences in sensor types result in varying image processing algorithms, with the RGGB and RYYB sensors requiring distinct demosaic methods, white balance (WB) corrections, and color calibration. The RYYB sensor, being more sensitive to light and affected by color lens shading (caused by small focal length and wide lens angle), also demands different tone-mapping techniques. Furthermore, the color gamut and saturation levels differ significantly between RGB and RYB, especially under low-light conditions. These variations in sensor behavior introduce a substantial domain gap between images captured by the two cameras, which is ideal for evaluating color-matching methods. The dataset thus simulates the real-world challenge of color matching between cameras.

\begin{figure}[!th]
  \centering
  \includegraphics[width=1.\linewidth]{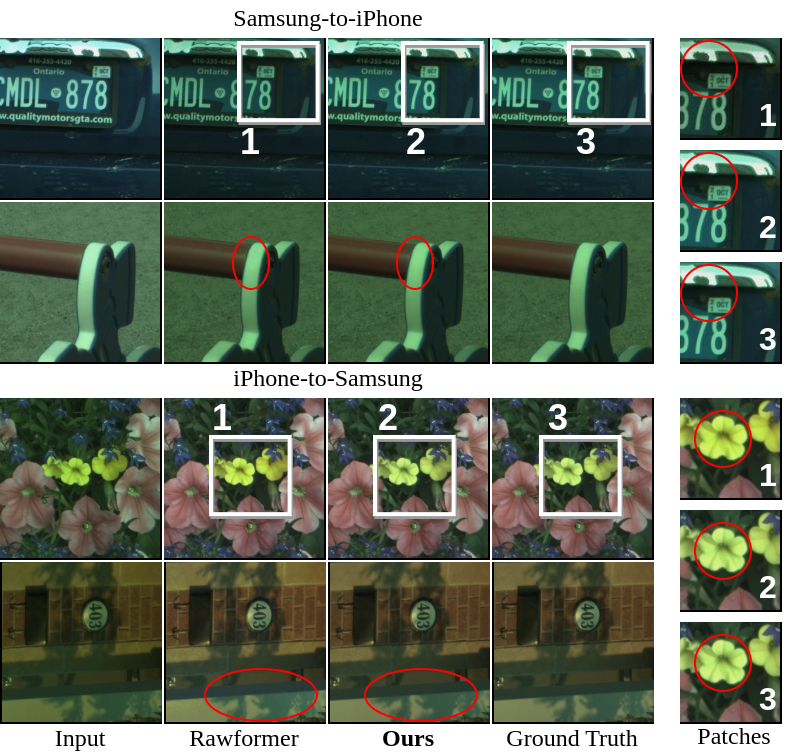}
  \vspace{-3mm}
  \caption{Qualitative results of raw translation on the Raw-to-Raw dataset~\cite{afifi2021semi}. Shown are images captured by Samsung S9 and iPhone X. We show the input raw patch from the corresponding camera and the corresponding ground-truth raw patch from the other camera, along with the results by RawFormer~\cite{perevozchikov2024rawformer}.}
  \label{fig:raw2}
\end{figure}

Our dataset, spanning four years of data collection, includes images captured at over 20 locations across four countries. The images are well-aligned with minimal matching errors (see Fig.~\ref{fig:dataset}), and we provide detailed annotations for various scenes (see Table~\ref{tab:our-dataset}). 
Additionally, our dataset includes keypoint clouds generated using the SURF detector~\cite{bay2006surf} for each image pair, along with binary matching masks to exclude misaligned regions. While our proposed method performs well even with unaligned data, this additional information is valuable for future research. To ensure ease of access and usability, we also offer a command-line interface (CLI) toolkit for processing, preparing, and cropping the dataset into smaller sections, making it more convenient for use in a variety of tasks. Our dataset and the CLI toolkit will be made available upon acceptance.

\begin{table}[h]
   \centering
   \caption{Scene and image classes in our dataset.}
   \resizebox{\linewidth}{!}
   {
   \begin{tabular}{@{}l|r|l|r@{}}
     \toprule
     Scene class        & Count & Image class & Count  \\
     \midrule
     Indoor            & 114 & City & 324 \\
     Outdoor/spring    & 109 & Countryside & 164 \\
     Outdoor/summer    & 558 & Forest & 49 \\
     Outdoor/autumn    & 175 & Mountains & 166 \\
     Outdoor/winter    & 97 & Seaside & 161 \\
     Low-light       & 107 & Sunset & 92 \\
     \textbf{Total}    & 1260 & Difficult illumination & 21 \\
     \bottomrule
   \end{tabular}
   }
   \label{tab:our-dataset}
\end{table}

\begin{figure}[!ht]
  \centering
  \includegraphics[width=1.\linewidth]{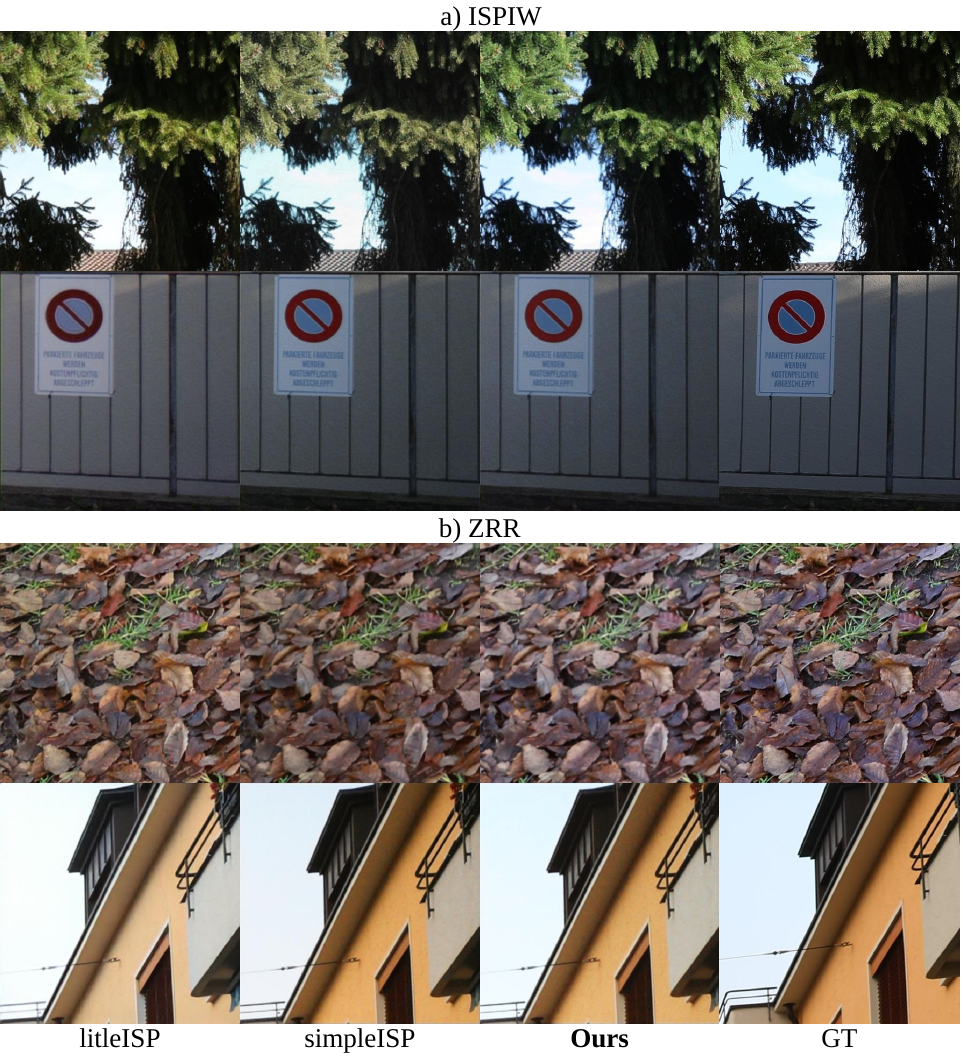}
  \vspace{-3mm}
  \caption{Qualitative comparison of raw-to-sRGB rendering on a) the ISPIW~\cite{shekhar2022transform} and b) the Zurich raw-to-sRGB~\cite{ignatov2020replacing} datasets. Our cmKAN method achieves the most accurate color mapping than LiteISP~\cite{zhang2021learning} and SimpleISP~\cite{elezabi2024simple}. Best viewed in the electronic version.}
  \label{fig:isp2}
\end{figure}
\section{Additional Results} \label{sec:extra-results}

\begin{figure*}[!ht]
  \centering
  \includegraphics[width=1.\linewidth]{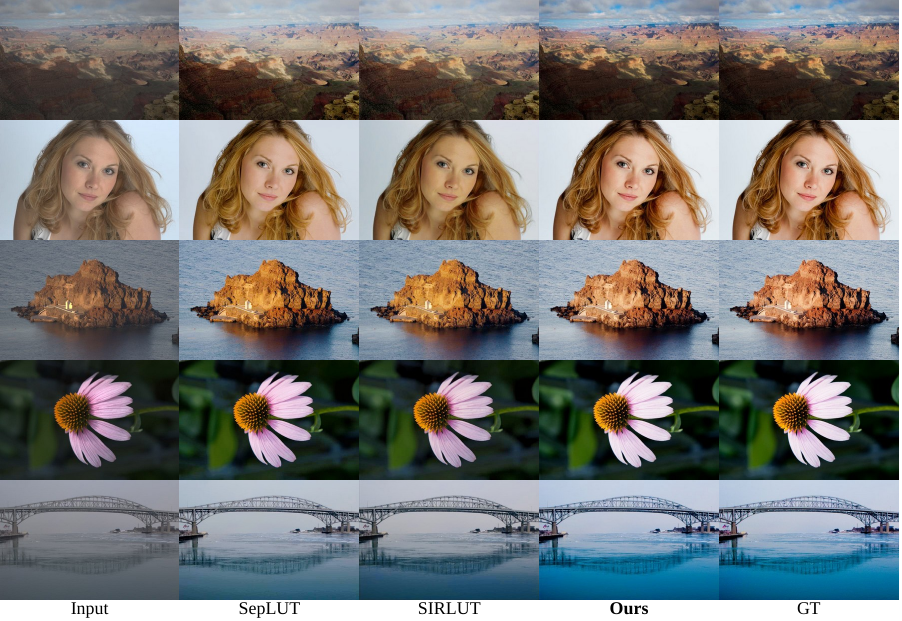}
  \vspace{-3mm}
  \caption{Qualitative results of sRGB-to-sRGB translation on the Adobe FiveK dataset~\cite{AN7}. Our method demonstrates SoTA results compared to other methods (SepLUT~\cite{yang2022seplut} and SIRLUT~\cite{li2024sirlut}).}
  \label{fig:five_k}
\end{figure*}

\subsection{Raw-to-Raw Mapping}

In this section, we provide additional qualitative and quantitative results for \textit{unsupervised} raw-to-raw mapping. Using the NUS dataset~\cite{cheng2014illuminant}, we performed mapping tasks between Nikon D5200 and Canon EOS 600D DSLR cameras, following the evaluation protocol described in~\cite{afifi2021semi, perevozchikov2024rawformer}. These results are summarized in Table~\ref{tab:raw}. As shown, our method achieves state-of-the-art performance across all metrics. Our method improves visual fidelity while preserving structural consistency, as confirmed by the superior SSIM and $\Delta E$ values. In addition, Fig.~\ref{fig:raw2} shows visual comparisons.

\begin{table}[t]
  \centering
    \caption{Results for \textit{unsupervised} raw-to-raw mapping using NUS dataset~\cite{cheng2014illuminant}. The best results are highlighted in yellow. \label{tab:nus}}
    \vspace{-2mm}
  \resizebox{1.0\linewidth}{!}
  {
  \begin{tabular}{@{}l|ccc|ccc@{}}
    \toprule
    Method & PSNR & SSIM & $\Delta E$ & PSNR & SSIM & $\Delta E$\\
    & \multicolumn{3}{c|}{\colorbox{supervised}{Canon-to-Nikon}} & \multicolumn{3}{c}{\colorbox{unsipervised}{Nikon-to-Canon}}\\
    \midrule
    SSRM~\cite{afifi2021semi}    & 32.36 & 0.93 & 6.21 & 30.81 & 0.93 & 5.95 \\
UVCGANv2~\cite{torbunov2023uvcgan2}  & 37.11 & 0.96 & 4.34 & 37.29 & 0.96 & 4.28 \\
    RawFormer~\cite{perevozchikov2024rawformer}& 41.89 & \colorbox{best}{0.98} & 2.04 & 41.37 & \colorbox{best}{0.98} & 2.53 \\
    cmKAN     & \colorbox{best}{41.93} & \colorbox{best}{0.98} & \colorbox{best}{1.61} & \colorbox{best}{41.62} & \colorbox{best}{0.98} & \colorbox{best}{1.54} \\
    \bottomrule
  \end{tabular}
  }  \vspace{-2mm}
  \label{tab:raw}
\end{table}

\subsection{Raw-to-sRGB Mapping}

In addition to the reported results in the main paper, we evaluate our method on the \textit{supervised} raw-to-sRGB mapping task using the ISPIW dataset~\cite{shekhar2022transform}. This dataset comprises raw images captured with a Huawei Mate 30 Pro smartphone and their corresponding sRGB reference images taken with a Canon 5D Mark IV camera. For our experiments, we utilize all 192 full-resolution images available, splitting them into 90\% for training and 10\% for testing. This setup yields 5,152 patch pairs for training and 572 for testing. The results are presented in Table~\ref{tab:aim}. Our method demonstrates superior performance, achieving state-of-the-art results across all evaluated metrics. In Fig.~\ref{fig:isp2}, we show qualitative results on the ZRR~\cite{ignatov2020replacing} and the ISPIW~\cite{shekhar2022transform} datasets.

\begin{table}[h]
  \centering
    \caption{Results for \textit{supervised} raw-to-sRGB mapping on the ISPIW raw-to-sRGB dataset~\cite{shekhar2022transform}.}
      \label{tab:aim}
          \vspace{-2mm}
  \resizebox{0.7\linewidth}{!}
  {
  \begin{tabular}{@{}l|ccc@{}}
    \toprule
    Method & PSNR & SSIM & $\Delta E$ \\
    \midrule
    MW-ISP~\cite{ignatov2020aim}  & 21.90 &  0.81 & 7.03\\
    LiteISP~\cite{zhang2021learning} & 22.14 &  0.81 & 6.31\\
    MicroISP~\cite{ignatov2022microisp} & 20.70 & 0.77 & 6.92\\
    SimpleISP~\cite{elezabi2024simple} & 23.67 & 0.82 & 5.91\\
    cmKAN   & \colorbox{best}{24.22} & \colorbox{best}{0.83} &  \colorbox{best}{5.29}\\
    \bottomrule
  \end{tabular}
  \vspace{-5mm}
  }
\end{table}

\subsection{sRGB-to-sRGB Mapping}

\begin{figure}[!ht]
  \centering
  \includegraphics[width=1.\linewidth]{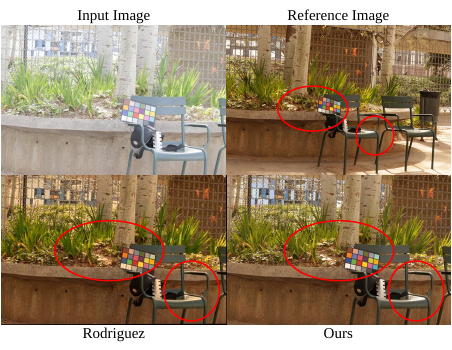}
  \vspace{-3mm}
  \caption{Qualitative comparison of raw-to-sRGB rendering on R.G. Rodriguez \textit{et al.}~\cite{AN2} dataset for 19th image. Our cmKAN-Light method achieves the most accurate color mapping compared with the original R.G. Rodriguez \textit{et al.}~\cite{AN2} method. Best viewed in the electronic version.}
  \label{fig:jav}
\end{figure}

To further evaluate our method, we provide additional results for \textit{supervised} sRGB-to-sRGB mapping using the PPR10K dataset~\cite{AN8} and \textit{paired-based optimization} on the R.G. Rodriguez \textit{et al.}~\cite{AN2} dataset.

\begin{figure*}[!ht]
  \centering
  \includegraphics[width=1.\linewidth]{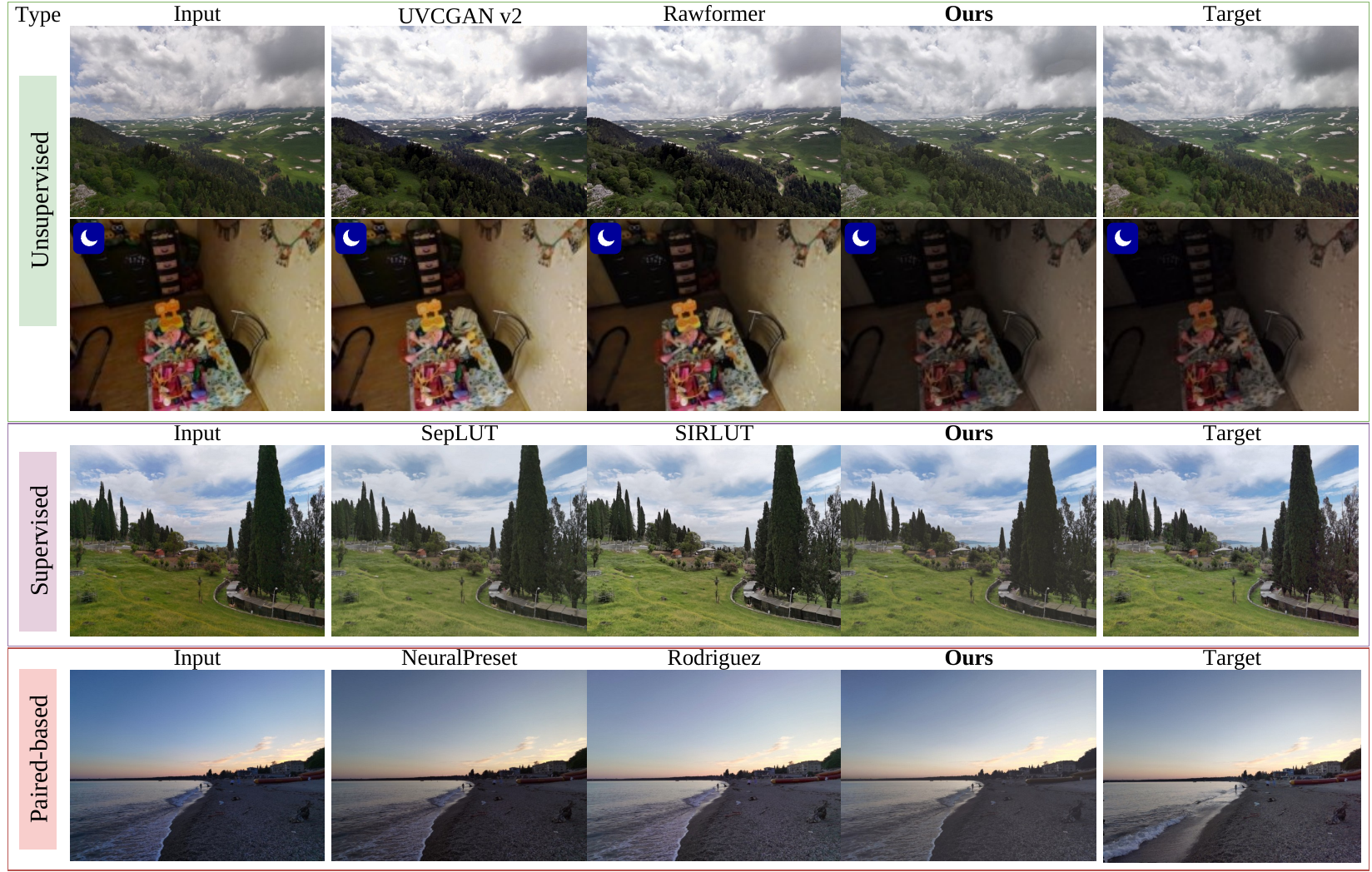}
  \vspace{-3mm}
  \caption{Additional results of sRGB-to-sRGB mapping on the proposed dataset. In (a), we show three scenarios using our method: unsupervised learning (green), supervised learning (purple), and paired-based optimization (red). Our method demonstrates superior color matching compared to other methods (UVCGAN v2~\cite{torbunov2023uvcgan2}, Rawformer~\cite{perevozchikov2024rawformer}, SepLUT~\cite{yang2022seplut}, SIRLUT~\cite{li2024sirlut}, NeuralPreset~\cite{ke2023neural}, and Rodriguez \textit{et al}.'s method~\cite{AN2}), even under low-light conditions (marked with a moon icon).}
  \label{fig:samara2}
\end{figure*}

The PPR10K dataset consists of 10,000 raw images rendered by three expert photographers (Experts A, B, and C). In this experiment, we used 7,000 images for training and 3,000 for testing. Table~\ref{tab:cm-ppr10k} summarizes our results, where our method (cmKAN) achieves state-of-the-art performance across all metrics, significantly outperforming existing approaches. 

\begin{table}[!h]
  \centering
  \caption{Additional results on R.G. Rodriguez \textit{et al.}~\cite{AN2} dataset. We compare the results of \textit{paired-based inference}.}
  \resizebox{0.8\linewidth}{!}
  {
  \begin{tabular}{@{}l|rrr@{}}
    \toprule
    Method        & PSNR & SSIM & $\Delta E$ \\
    \midrule
    R.G. Rodriguez \textit{et al.}~\cite{AN2} & 26.91 & 0.84 & 3.26\\
    cmKAN-Light    & \colorbox{best}{28.21} & \colorbox{best}{0.86} & \colorbox{best}{3.07}\\
    \bottomrule
  \end{tabular}
  }
  \label{tab:cm-javier}
\end{table}

The R.G. Rodriguez \textit{et al.}~\cite{AN2} dataset contains 35 images captured using two camera models (Nikon D3100 and Canon EOS80D). Due to the limited dataset size, we performed \textit{paired-based optimization} for evaluation. Our results, detailed in Table~\ref{tab:cm-javier}, show that our lightweight model variant (cmKAN-Light) achieves the best scores in PSNR, SSIM, and $\Delta E$, further validating the versatility and adaptability of our approach. 

Additional qualitative results of sRGB-to-sRGB mapping are shown in Fig.~\ref{fig:five_k}, \ref{fig:jav} and ~\ref{fig:samara2}, showcasing our method's ability to handle diverse sRGB-to-sRGB mapping tasks effectively.

\begin{table}[h]
  \centering
  \caption{Results of \textit{supervised} sRGB-to-sRGB mapping on the PPR10K dataset~\cite{AN8}. We report training results using experts A, B, and C as ground-truth targets.}
      \vspace{-2mm}
  \resizebox{1.0\linewidth}{!}
  {
  \begin{tabular}{@{}l|c|c|c@{}}
    \toprule
    Method & PSNR / $\Delta E$ & PSNR / $\Delta E$ & PSNR / $\Delta E$\\
    \midrule
    & \multicolumn{1}{c|}{\colorbox{unsipervised}{Expert A}} & \multicolumn{1}{c|}{\colorbox{supervised}{Expert B}} & \multicolumn{1}{c}{\colorbox{paired}{Expert C}}\\
    SepLUT~\cite{yang2022seplut}        & 26.28 / 6.59 & 25.23 / 7.49 & 25.59 / 7.51\\
    LYT-Net~\cite{brateanu2024lyt}       & 26.10 / 7.03 & 23.93 / 9.21 & 23.93 / 9.21\\
    SIRLUT~\cite{li2024sirlut}       & 28.31 / 5.65 & 27.67 / 5.89 & 27.79 / 6.13\\
    cmKAN  & \colorbox{best}{32.49} / \colorbox{best}{2.21} & \colorbox{best}{32.01} / \colorbox{best}{2.35} & \colorbox{best}{32.27} / \colorbox{best}{2.77}\\     \bottomrule
  \end{tabular}
  }
  \label{tab:cm-ppr10k}
\end{table}

\section{User study experiment} \label{sec:user-study}

\begin{figure}[!th]
  \centering
  \includegraphics[width=1.\linewidth]{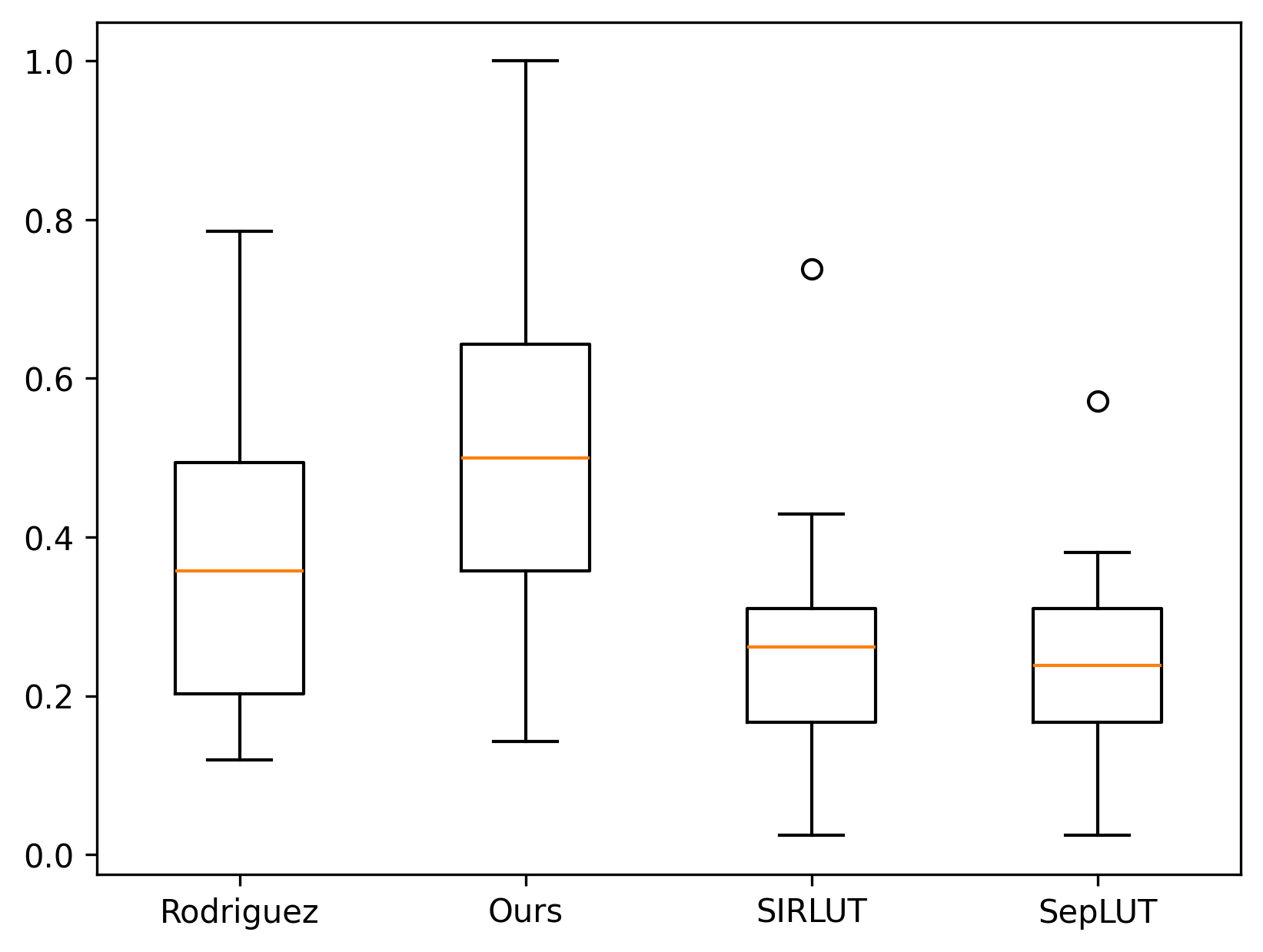}
  \vspace{-3mm}
  \caption{Normalized win count aggregated on all images.}
  \label{fig:hs}
\end{figure}

\begin{figure}[t]
  \centering
  \includegraphics[width=1.\linewidth]{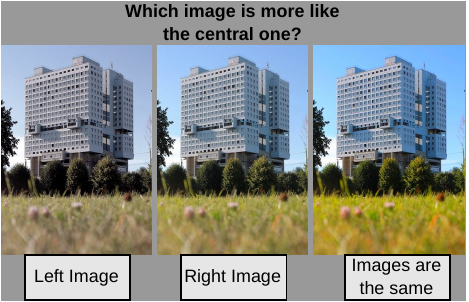}
  \vspace{-3mm}
  \caption{Illustration of a voting interface in the ``Toloka" service.}
  \label{fig:sh_ui}
\end{figure}


In addition to the quantitative and qualitative comparisons presented in the main paper and this supplementary material, we conducted a user study to validate our method from a human-guidance perspective.

We conducted the user study using the `Toloka' crowdsourcing platform. Thirty scenes from a test sample of the Adobe FiveK dataset~\cite{AN7} were processed with four different methods: 1) SepLUT~\cite{yang2022seplut}, 2) SIRLUT~\cite{li2024sirlut}, 3) R.G. Rodriguez \textit{et al.}~\cite{AN2}, and 4) ours.
Each scene generates $6$ combinations of style pairs, resulting in 6 $\times$ 30 $=$ 180 pairs for style comparison. 
It is important to note that the comparisons also included ``filtering'' pairs of identical images to exclude unscrupulous participants. On each Toloka webpage, five random image pairs were displayed, with the reference image placed in the center and the other images on the sides. For each, participants were asked to answer the question: `Which image was more similar to the center image?' The images were displayed on a 50\% gray background, as color comparisons should be conducted on a neutral background to eliminate bias~\cite{mantiuk2012comparison}. Additionally, we ensured that the space between the two images contained no other information or controls, to prevent distractions for the participants. To facilitate the comparison process, a gap was maintained between the images. For each pair, participants were presented with three response options; an example of the interface is shown in Fig.~\ref{fig:sh_ui}):

\begin{itemize}
    \item The right image is more like the center image
    \item The left image is more similar to the center image
    \item Both images are the same
\end{itemize}

A total of 431 participants took part in the experiments. Votes were not counted if participants marked images as similar. Additionally, if a participant failed the `filter' pair, the entire set of votes from that participant was discarded, and they were banned from the study.

The evaluation methodology for obtaining mean opinion scores, shown in Fig.~\ref{fig:hs}, uses a modified Bradley-Terry model~\cite{ershov2024reliability} to rank image versions based on pairwise comparisons from participant ratings, simplifying the process by assuming negligible observer dependence and averaging favorable votes. \textit{Our non-reference method outperforms existing methods by a factor of two}, which helps justify the significance of the results. Additionally, our method outperforms paired-based solutions~\cite{AN2} by$\times$1.5. The results of this study demonstrate that our proposed cmKAN significantly outperforms other methods in color matching.



{
    \small
    \bibliographystyle{ieeenat_fullname}
    \bibliography{main}

\begin{thebibliography}{79}
\providecommand{\natexlab}[1]{#1}
\providecommand{\url}[1]{\texttt{#1}}
\expandafter\ifx\csname urlstyle\endcsname\relax
  \providecommand{\doi}[1]{doi: #1}\else
  \providecommand{\doi}{doi: \begingroup \urlstyle{rm}\Url}\fi

\bibitem[Afifi and Abuolaim(2021)]{afifi2021semi}
Mahmoud Afifi and Abdullah Abuolaim.
\newblock Semi-supervised raw-to-raw mapping.
\newblock In \emph{BMVC}, 2021.

\bibitem[Afifi and Brown(2020)]{afifi2020deep}
Mahmoud Afifi and Michael~S Brown.
\newblock Deep white-balance editing.
\newblock In \emph{CVPR}, 2020.

\bibitem[Afifi et~al.(2019{\natexlab{a}})Afifi, Price, Cohen, and Brown]{afifi2019color}
Mahmoud Afifi, Brian Price, Scott Cohen, and Michael~S Brown.
\newblock When color constancy goes wrong: Correcting improperly white-balanced images.
\newblock In \emph{CVPR}, 2019{\natexlab{a}}.

\bibitem[Afifi et~al.(2019{\natexlab{b}})Afifi, Price, Cohen, and Brown]{afifi2019image}
Mahmoud Afifi, Brian~L Price, Scott Cohen, and Michael~S Brown.
\newblock Image recoloring based on object color distributions.
\newblock In \emph{Eurographics (Short Papers)}, 2019{\natexlab{b}}.

\bibitem[Afifi et~al.(2021{\natexlab{a}})Afifi, Brubaker, and Brown]{afifi2021histogan}
Mahmoud Afifi, Marcus~A Brubaker, and Michael~S Brown.
\newblock Histogan: Controlling colors of gan-generated and real images via color histograms.
\newblock In \emph{CVPR}, 2021{\natexlab{a}}.

\bibitem[Afifi et~al.(2021{\natexlab{b}})Afifi, Derpanis, Ommer, and Brown]{afifi2021learning}
Mahmoud Afifi, Konstantinos~G Derpanis, Bjorn Ommer, and Michael~S Brown.
\newblock Learning multi-scale photo exposure correction.
\newblock In \emph{CVPR}, 2021{\natexlab{b}}.

\bibitem[Alzayer et~al.(2023)Alzayer, Abuolaim, Chan, Yang, Lou, Huang, and Kar]{alzayer2023dc2}
Hadi Alzayer, Abdullah Abuolaim, Leung~Chun Chan, Yang Yang, Ying~Chen Lou, Jia-Bin Huang, and Abhishek Kar.
\newblock {DC}2: Dual-camera defocus control by learning to refocus.
\newblock In \emph{CVPR}, 2023.

\bibitem[Bay et~al.(2006)Bay, Tuytelaars, and Van~Gool]{bay2006surf}
Herbert Bay, Tinne Tuytelaars, and Luc Van~Gool.
\newblock {SURF}: Speeded up robust features.
\newblock In \emph{ECCV}, 2006.

\bibitem[Bianco et~al.(2012)Bianco, Bruna, Naccari, and Schettini]{AN9}
Simone Bianco, Arcangelo Bruna, Filippo Naccari, and Raimondo Schettini.
\newblock Color space transformations for digital photography exploiting information about the illuminant estimation process.
\newblock \emph{J. Opt. Soc. Am. A}, 29\penalty0 (3):\penalty0 374--384, 2012.

\bibitem[Brateanu et~al.(2024)Brateanu, Balmez, Avram, and Orhei]{brateanu2024lyt}
Alexandru Brateanu, Raul Balmez, Adrian Avram, and CC Orhei.
\newblock {LYT-NET}: Lightweight {YUV} transformer-based network for low-light image enhancement.
\newblock \emph{arXiv preprint arXiv:2401.15204}, 2024.

\bibitem[Brateanu et~al.(2025)Brateanu, Balmez, Orhei, Ancuti, and Ancuti]{brateanu2025enhancing}
Alexandru Brateanu, Raul Balmez, Ciprian Orhei, Cosmin Ancuti, and Codruta Ancuti.
\newblock Enhancing low-light images with kolmogorov--arnold networks in transformer attention.
\newblock \emph{Sensors}, 25\penalty0 (2):\penalty0 327, 2025.

\bibitem[Brown(2023)]{brown2023color}
MichaelS Brown.
\newblock Color processing for digital cameras.
\newblock \emph{Fundamentals and Applications of Colour Engineering}, pages 81--98, 2023.

\bibitem[Bychkovsky et~al.(2011)Bychkovsky, Paris, Chan, and Durand]{AN7}
Vladimir Bychkovsky, Sylvain Paris, Eric Chan, and Fr{\'e}do Durand.
\newblock Learning photographic global tonal adjustment with a database of input/output image pairs.
\newblock In \emph{CVPR}, 2011.

\bibitem[Cai et~al.(2022)Cai, Lin, Lin, Wang, Zhang, Pfister, Timofte, and Van~Gool]{mst++}
Yuanhao Cai, Jing Lin, Zudi Lin, Haoqian Wang, Yulun Zhang, Hanspeter Pfister, Radu Timofte, and Luc Van~Gool.
\newblock {MST}++: Multi-stage spectral-wise transformer for efficient spectral reconstruction.
\newblock In \emph{CVPR}, 2022.

\bibitem[Chang et~al.(2015)Chang, Fried, Liu, DiVerdi, and Finkelstein]{chang2015palette}
Huiwen Chang, Ohad Fried, Yiming Liu, Stephen DiVerdi, and Adam Finkelstein.
\newblock Palette-based photo recoloring.
\newblock \emph{ACM Transactions on Graphics}, 34\penalty0 (4):\penalty0 139--1, 2015.

\bibitem[Chen et~al.(2018)Chen, Wang, Kao, and Chuang]{chen2018deep}
Yu-Sheng Chen, Yu-Ching Wang, Man-Hsin Kao, and Yung-Yu Chuang.
\newblock Deep photo enhancer: {U}npaired learning for image enhancement from photographs with {GAN}s.
\newblock In \emph{CVPR}, 2018.

\bibitem[Cheng et~al.(2014)Cheng, Prasad, and Brown]{cheng2014illuminant}
Dongliang Cheng, Dilip~K Prasad, and Michael~S Brown.
\newblock Illuminant estimation for color constancy: why spatial-domain methods work and the role of the color distribution.
\newblock \emph{JOSA A}, 31\penalty0 (5):\penalty0 1049--1058, 2014.

\bibitem[Conde et~al.(2024)Conde, Vazquez-Corral, Brown, and Timofte]{conde2024nilut}
Marcos~V Conde, Javier Vazquez-Corral, Michael~S Brown, and Radu Timofte.
\newblock {NILUT}: Conditional neural implicit 3{D} lookup tables for image enhancement.
\newblock In \emph{AAAI}, 2024.

\bibitem[Delbracio et~al.(2021)Delbracio, Kelly, Brown, and Milanfar]{delbracio2021mobile}
Mauricio Delbracio, Damien Kelly, Michael~S Brown, and Peyman Milanfar.
\newblock Mobile computational photography: A tour.
\newblock \emph{Annual review of vision science}, 7\penalty0 (1):\penalty0 571--604, 2021.

\bibitem[Ding et~al.(2024)Ding, Li, Yang, Li, and Gong]{ding2024regional}
Zhicheng Ding, Panfeng Li, Qikai Yang, Siyang Li, and Qingtian Gong.
\newblock Regional style and color transfer.
\newblock In \emph{CVIDL}, 2024.

\bibitem[Elezabi et~al.(2024)Elezabi, Conde, and Timofte]{elezabi2024simple}
Omar Elezabi, Marcos~V Conde, and Radu Timofte.
\newblock Simple image signal processing using global context guidance.
\newblock In \emph{ICIP}, 2024.

\bibitem[Ershov et~al.(2024)Ershov, Panshin, Ermakov, Banic, Savchik, and Bianco]{ershov2024reliability}
Egor Ershov, Artyom Panshin, Ivan Ermakov, Nikola Banic, Alex Savchik, and Simone Bianco.
\newblock Reliability and stability of mean opinion score for image aesthetic quality assessment obtained through crowdsourcing.
\newblock \emph{Proceedings Copyright}, 365:\penalty0 372, 2024.

\bibitem[Fairchild et~al.(2008)Fairchild, Wyble, and Johnson]{fairchild2008matching}
Mark~D Fairchild, David~R Wyble, and Garrett~M Johnson.
\newblock Matching image color from different cameras.
\newblock In \emph{Image Quality and System Performance V}, 2008.

\bibitem[Faridul et~al.(2014)Faridul, Pouli, Chamaret, Stauder, Tr{\'e}meau, Reinhard, et~al.]{faridul2014survey}
Hasan~Sheikh Faridul, Tania Pouli, Christel Chamaret, J{\"u}rgen Stauder, Alain Tr{\'e}meau, Erik Reinhard, et~al.
\newblock A survey of color mapping and its applications.
\newblock \emph{Eurographics (State of the Art Reports)}, 3\penalty0 (2):\penalty0 1, 2014.

\bibitem[Finlayson et~al.(2017)Finlayson, Gong, and Fisher]{finlayson2017color}
Graham Finlayson, Han Gong, and Robert~B Fisher.
\newblock Color homography: theory and applications.
\newblock \emph{IEEE Transactions on Pattern Analysis and Machine Intelligence}, 41\penalty0 (1):\penalty0 20--33, 2017.

\bibitem[Finlayson et~al.(2015)Finlayson, Mackiewicz, and Hurlbert]{finlayson2015color}
Graham~D Finlayson, Michal Mackiewicz, and Anya Hurlbert.
\newblock Color correction using root-polynomial regression.
\newblock \emph{IEEE Transactions on Image Processing}, 24\penalty0 (5):\penalty0 1460--1470, 2015.

\bibitem[Gil~Rodríguez et~al.(2020)Gil~Rodríguez, Vazquez-Corral, and Bertalmío]{AN2}
Raquel Gil~Rodríguez, Javier Vazquez-Corral, and Marcelo Bertalmío.
\newblock Color matching images with unknown non-linear encodings.
\newblock \emph{IEEE Transactions on Image Processing}, 29:\penalty0 4435--4444, 2020.

\bibitem[{Gil Rodríguez} et~al.(2024){Gil Rodríguez}, Vazquez-Corral, Bertalmío, and Finlayson]{AN1}
Raquel {Gil Rodríguez}, Javier Vazquez-Corral, Marcelo Bertalmío, and Graham~D. Finlayson.
\newblock Color matching in the wild.
\newblock \emph{Pattern Recognition}, 154:\penalty0 110575, 2024.

\bibitem[Ha et~al.(2016)Ha, Dai, and Le]{ha2016hypernetworks}
David Ha, Andrew Dai, and Quoc~V Le.
\newblock Hypernetworks.
\newblock \emph{arXiv preprint arXiv:1609.09106}, 2016.

\bibitem[Heide et~al.(2014)Heide, Steinberger, Tsai, Rouf, Paj{\k{a}}k, Reddy, Gallo, Liu, Heidrich, Egiazarian, et~al.]{heide2014flexisp}
Felix Heide, Markus Steinberger, Yun-Ta Tsai, Mushfiqur Rouf, Dawid Paj{\k{a}}k, Dikpal Reddy, Orazio Gallo, Jing Liu, Wolfgang Heidrich, Karen Egiazarian, et~al.
\newblock Flexisp: A flexible camera image processing framework.
\newblock \emph{ACM Transactions on Graphics}, 33\penalty0 (6):\penalty0 1--13, 2014.

\bibitem[Hong et~al.(2001)Hong, Luo, and Rhodes]{hong2001study}
Guowei Hong, M~Ronnier Luo, and Peter~A Rhodes.
\newblock A study of digital camera colorimetric characterization based on polynomial modeling.
\newblock \emph{Color Research \& Application}, 26\penalty0 (1):\penalty0 76--84, 2001.

\bibitem[Ignatov et~al.(2020{\natexlab{a}})Ignatov, Timofte, Zhang, Liu, Wang, Zuo, Zhang, Zhang, Peng, Ren, et~al.]{ignatov2020aim}
Andrey Ignatov, Radu Timofte, Zhilu Zhang, Ming Liu, Haolin Wang, Wangmeng Zuo, Jiawei Zhang, Ruimao Zhang, Zhanglin Peng, Sijie Ren, et~al.
\newblock {AIM} 2020 challenge on learned image signal processing pipeline.
\newblock In \emph{ECCVW}, 2020{\natexlab{a}}.

\bibitem[Ignatov et~al.(2020{\natexlab{b}})Ignatov, Van~Gool, and Timofte]{ignatov2020replacing}
Andrey Ignatov, Luc Van~Gool, and Radu Timofte.
\newblock Replacing mobile camera {ISP} with a single deep learning model.
\newblock In \emph{CVPRW}, 2020{\natexlab{b}}.

\bibitem[Ignatov et~al.(2022)Ignatov, Sycheva, Timofte, Tseng, Xu, Yu, Chiang, Kuo, Chen, Cheng, et~al.]{ignatov2022microisp}
Andrey Ignatov, Anastasia Sycheva, Radu Timofte, Yu Tseng, Yu-Syuan Xu, Po-Hsiang Yu, Cheng-Ming Chiang, Hsien-Kai Kuo, Min-Hung Chen, Chia-Ming Cheng, et~al.
\newblock Micro{ISP}: processing 32mp photos on mobile devices with deep learning.
\newblock In \emph{ECCV}, 2022.

\bibitem[Karaimer and Brown(2016)]{karaimer2016software}
Hakki~Can Karaimer and Michael~S Brown.
\newblock A software platform for manipulating the camera imaging pipeline.
\newblock In \emph{ECCV}, 2016.

\bibitem[Ke et~al.(2023)Ke, Liu, Zhu, Zhao, and Lau]{ke2023neural}
Zhanghan Ke, Yuhao Liu, Lei Zhu, Nanxuan Zhao, and Rynson~WH Lau.
\newblock Neural preset for color style transfer.
\newblock In \emph{CVPR}, 2023.

\bibitem[Kingma(2014)]{kingma2014adam}
Diederik~P Kingma.
\newblock Adam: A method for stochastic optimization.
\newblock \emph{arXiv preprint arXiv:1412.6980}, 2014.

\bibitem[Kolmogorov(1957)]{kolmogorov:superposition}
A.~K. Kolmogorov.
\newblock On the representation of continuous functions of several variables by superposition of continuous functions of one variable and addition.
\newblock \emph{Doklady Akademii Nauk SSSR}, 114:\penalty0 369--373, 1957.

\bibitem[Kucuk et~al.(2022)Kucuk, Finlayson, Mantiuk, and Ashraf]{AN3}
Abdullah Kucuk, Graham Finlayson, Rafal Mantiuk, and Maliha Ashraf.
\newblock Comparison of regression methods and neural networks for colour corrections.
\newblock In \emph{London Imaging Meeting}, pages 74--79, 2022.

\bibitem[Kucuk et~al.(2023)Kucuk, Finlayson, Mantiuk, and Ashraf]{AN4}
Abdullah Kucuk, Graham~D. Finlayson, Rafal Mantiuk, and Maliha Ashraf.
\newblock Performance comparison of classical methods and neural networks for colour correction.
\newblock \emph{Journal of Imaging}, 9\penalty0 (10), 2023.

\bibitem[Lai et~al.(2022)Lai, Shih, Chu, Wu, Tsai, Krainin, Sun, and Liang]{lai2022face}
Wei-Sheng Lai, Yichang Shih, Lun-Cheng Chu, Xiaotong Wu, Sung-Fang Tsai, Michael Krainin, Deqing Sun, and Chia-Kai Liang.
\newblock Face deblurring using dual camera fusion on mobile phones.
\newblock \emph{ACM Transactions on Graphics}, 41\penalty0 (4):\penalty0 1--16, 2022.

\bibitem[Le et~al.(2023)Le, Price, Cohen, and Brown]{le2023gamutmlp}
Hoang~M Le, Brian Price, Scott Cohen, and Michael~S Brown.
\newblock Gamut{MLP}: A lightweight mlp for color loss recovery.
\newblock In \emph{CVPR}, 2023.

\bibitem[Li et~al.(2024)Li, Li, Li, Wang, Guo, and Jiang]{li2024sirlut}
Kaijiang Li, Hao Li, Haining Li, Peisen Wang, Chunyi Guo, and Wenfeng Jiang.
\newblock {SIRLUT}: Simulated infrared fusion guided image-adaptive 3{D} lookup tables for lightweight image enhancement.
\newblock In \emph{ACM MM}, 2024.

\bibitem[Liang et~al.(2021)Liang, Zeng, Cui, Xie, and Zhang]{AN8}
Jie Liang, Hui Zeng, Miaomiao Cui, Xuansong Xie, and Lei Zhang.
\newblock {PPR10K}: A large-scale portrait photo retouching dataset with human-region mask and group-level consistency.
\newblock In \emph{CVPR}, 2021.

\bibitem[Liu et~al.(2023)Liu, Yang, Fu, and Qian]{liu20234d}
Chengxu Liu, Huan Yang, Jianlong Fu, and Xueming Qian.
\newblock 4{D} {LUT}: learnable context-aware 4{D} lookup table for image enhancement.
\newblock \emph{IEEE Transactions on Image Processing}, 32:\penalty0 4742--4756, 2023.

\bibitem[Liu et~al.(2024)Liu, Wang, Vaidya, Ruehle, Halverson, Solja{\v{c}}i{\'c}, Hou, and Tegmark]{AN6}
Ziming Liu, Yixuan Wang, Sachin Vaidya, Fabian Ruehle, James Halverson, Marin Solja{\v{c}}i{\'c}, Thomas~Y Hou, and Max Tegmark.
\newblock Kan: Kolmogorov-arnold networks.
\newblock \emph{arXiv preprint arXiv:2404.19756}, 2024.

\bibitem[Lowe(2004)]{lowe2004distinctive}
David~G Lowe.
\newblock Distinctive image features from scale-invariant keypoints.
\newblock \emph{International journal of computer vision}, 60:\penalty0 91--110, 2004.

\bibitem[Lv et~al.(2024)Lv, Zhang, Geng, Wu, and Huang]{lv2024color}
Chenlei Lv, Dan Zhang, Shengling Geng, Zhongke Wu, and Hui Huang.
\newblock Color transfer for images: A survey.
\newblock \emph{ACM Transactions on Multimedia Computing, Communications and Applications}, 20\penalty0 (8):\penalty0 1--29, 2024.

\bibitem[Mantiuk et~al.(2012)Mantiuk, Tomaszewska, and Mantiuk]{mantiuk2012comparison}
Rafa{\l}~K Mantiuk, Anna Tomaszewska, and Rados{\l}aw Mantiuk.
\newblock {C}omparison of {F}our {S}ubjective {M}ethods for {I}mage {Q}uality {A}ssessment.
\newblock In \emph{Computer graphics forum}, pages 2478--2491. Wiley Online Library, 2012.

\bibitem[Menesatti et~al.(2012)Menesatti, Angelini, Pallottino, Antonucci, Aguzzi, and Costa]{menesatti2012rgb}
Paolo Menesatti, Claudio Angelini, Federico Pallottino, Francesca Antonucci, Jacopo Aguzzi, and Corrado Costa.
\newblock {RGB} color calibration for quantitative image analysis: The “3d thin-plate spline” warping approach.
\newblock \emph{Sensors}, 12\penalty0 (6):\penalty0 7063--7079, 2012.

\bibitem[Nakamura(2017)]{nakamura2017image}
Junichi Nakamura.
\newblock \emph{Image sensors and signal processing for digital still cameras}.
\newblock CRC press, 2017.

\bibitem[Nam et~al.(2022)Nam, Punnappurath, Brubaker, and Brown]{nam2022learning}
Seonghyeon Nam, Abhijith Punnappurath, Marcus~A Brubaker, and Michael~S Brown.
\newblock Learning s{RGB}-to-raw-{RGB} de-rendering with content-aware metadata.
\newblock In \emph{CVPR}, 2022.

\bibitem[Nguyen et~al.(2014{\natexlab{a}})Nguyen, Prasad, and Brown]{nguyen2014raw}
Rang Nguyen, Dilip~K Prasad, and Michael~S Brown.
\newblock Raw-to-raw: Mapping between image sensor color responses.
\newblock In \emph{CVPR}, 2014{\natexlab{a}}.

\bibitem[Nguyen et~al.(2014{\natexlab{b}})Nguyen, Kim, and Brown]{nguyen2014illuminant}
Rang~MH Nguyen, Seon~Joo Kim, and Michael~S Brown.
\newblock Illuminant aware gamut-based color transfer.
\newblock In \emph{Computer Graphics Forum}, 2014{\natexlab{b}}.

\bibitem[Nikonorov et~al.(2016)Nikonorov, Bibikov, Myasnikov, Yuzifovich, and Fursov]{nikonorov2016correcting}
Artem Nikonorov, Sergey Bibikov, Vladislav Myasnikov, Yuriy Yuzifovich, and Vladimir Fursov.
\newblock Correcting color and hyperspectral images with identification of distortion model.
\newblock \emph{Pattern Recognition Letters}, 83:\penalty0 178--187, 2016.

\bibitem[Perevozchikov et~al.(2024)Perevozchikov, Mehta, Afifi, and Timofte]{perevozchikov2024rawformer}
Georgy Perevozchikov, Nancy Mehta, Mahmoud Afifi, and Radu Timofte.
\newblock Rawformer: Unpaired raw-to-raw translation for learnable camera {ISP}s.
\newblock In \emph{ECCV}, 2024.

\bibitem[Pitie et~al.(2005)Pitie, Kokaram, and Dahyot]{pitie2005n}
Francois Pitie, Anil~C Kokaram, and Rozenn Dahyot.
\newblock N-dimensional probability density function transfer and its application to color transfer.
\newblock In \emph{Tenth IEEE International Conference on Computer Vision (ICCV'05) Volume 1}, pages 1434--1439, 2005.

\bibitem[Piti{\'e} et~al.(2007)Piti{\'e}, Kokaram, and Dahyot]{pitie2007automated}
Fran{\c{c}}ois Piti{\'e}, Anil~C Kokaram, and Rozenn Dahyot.
\newblock Automated colour grading using colour distribution transfer.
\newblock \emph{Computer Vision and Image Understanding}, 107\penalty0 (1-2):\penalty0 123--137, 2007.

\bibitem[Punnappurath and Brown(2021)]{punnappurath2021spatially}
Abhijith Punnappurath and Michael~S Brown.
\newblock Spatially aware metadata for raw reconstruction.
\newblock In \emph{WACV}, 2021.

\bibitem[Rabin et~al.(2014)Rabin, Ferradans, and Papadakis]{rabin2014adaptive}
Julien Rabin, Sira Ferradans, and Nicolas Papadakis.
\newblock Adaptive color transfer with relaxed optimal transport.
\newblock In \emph{ICIP}, 2014.

\bibitem[Reinhard et~al.(2001)Reinhard, Adhikhmin, Gooch, and Shirley]{reinhard2001color}
Erik Reinhard, Michael Adhikhmin, Bruce Gooch, and Peter Shirley.
\newblock Color transfer between images.
\newblock \emph{IEEE Computer graphics and applications}, 21\penalty0 (5):\penalty0 34--41, 2001.

\bibitem[Schwartz et~al.(2018)Schwartz, Giryes, and Bronstein]{schwartz2018deepisp}
Eli Schwartz, Raja Giryes, and Alex~M Bronstein.
\newblock Deep{ISP}: Toward learning an end-to-end image processing pipeline.
\newblock \emph{IEEE Transactions on Image Processing}, 28\penalty0 (2):\penalty0 912--923, 2018.

\bibitem[Shekhar~Tripathi et~al.(2022)Shekhar~Tripathi, Danelljan, Shukla, Timofte, and Van~Gool]{shekhar2022transform}
Ardhendu Shekhar~Tripathi, Martin Danelljan, Samarth Shukla, Radu Timofte, and Luc Van~Gool.
\newblock Transform your smartphone into a dslr camera: Learning the isp in the wild.
\newblock In \emph{European Conference on Computer Vision}, pages 625--641. Springer, 2022.

\bibitem[Suominen and Egiazarian(2024)]{suominen2024camera}
Joni Suominen and Karen Egiazarian.
\newblock Camera color correction using splines.
\newblock \emph{Electronic Imaging}, 36:\penalty0 1--6, 2024.

\bibitem[Tocci et~al.(2022)Tocci, Figorilli, Vasta, Violino, Pallottino, Ortenzi, and Costa]{tocci2022advantages}
Francesco Tocci, Simone Figorilli, Simone Vasta, Simona Violino, Federico Pallottino, Luciano Ortenzi, and Corrado Costa.
\newblock Advantages in using colour calibration for orthophoto reconstruction.
\newblock \emph{Sensors}, 22\penalty0 (17):\penalty0 6490, 2022.

\bibitem[Torbunov et~al.(2023)Torbunov, Huang, Yu, Huang, Yoo, Lin, Viren, and Ren]{torbunov2023uvcgan2}
Dmitrii Torbunov, Yi Huang, Haiwang Yu, Jin Huang, Shinjae Yoo, Meifeng Lin, Brett Viren, and Yihui Ren.
\newblock {UVCGAN} v2: An improved cycle-consistent {GAN} for unpaired image-to-image translation.
\newblock \emph{arXiv preprint arXiv:2303.16280}, 2023.

\bibitem[Tseng et~al.(2022)Tseng, Zhang, Jebe, Zhang, Xia, Fan, Heide, and Chen]{tseng2022neural}
Ethan Tseng, Yuxuan Zhang, Lars Jebe, Xuaner Zhang, Zhihao Xia, Yifei Fan, Felix Heide, and Jiawen Chen.
\newblock Neural photo-finishing.
\newblock \emph{ACM Transactions on Graphics}, 41\penalty0 (6):\penalty0 238--1, 2022.

\bibitem[Wang et~al.(2019)Wang, Zhang, Fu, Shen, Zheng, and Jia]{wang2019underexposed}
Ruixing Wang, Qing Zhang, Chi-Wing Fu, Xiaoyong Shen, Wei-Shi Zheng, and Jiaya Jia.
\newblock Underexposed photo enhancement using deep illumination estimation.
\newblock In \emph{CVPR}, 2019.

\bibitem[Wang et~al.(2004)Wang, Bovik, Sheikh, and Simoncelli]{wang2004image}
Zhou Wang, Alan~C Bovik, Hamid~R Sheikh, and Eero~P Simoncelli.
\newblock Image quality assessment: {F}rom error visibility to structural similarity.
\newblock \emph{IEEE Transactions on Image Processing}, 13\penalty0 (4):\penalty0 600--612, 2004.

\bibitem[Wu et~al.(2024)Wu, Zhang, Yang, and Zuo]{wu2025dual}
Renlong Wu, Zhilu Zhang, Yu Yang, and Wangmeng Zuo.
\newblock Dual-camera smooth zoom on mobile phones.
\newblock In \emph{ECCV}, 2024.

\bibitem[Wu et~al.(2023)Wu, Lai, Shih, Herrmann, Krainin, Sun, and Liang]{wu2023efficient}
Xiaotong Wu, Wei-Sheng Lai, Yichang Shih, Charles Herrmann, Michael Krainin, Deqing Sun, and Chia-Kai Liang.
\newblock Efficient hybrid zoom using camera fusion on mobile phones.
\newblock \emph{ACM Transactions on Graphics}, 42\penalty0 (6):\penalty0 1--12, 2023.

\bibitem[Xiao and Ma(2006)]{xiao2006color}
Xuezhong Xiao and Lizhuang Ma.
\newblock Color transfer in correlated color space.
\newblock In \emph{ACM international conference on Virtual reality continuum and its applications}, 2006.

\bibitem[Xing et~al.(2021)Xing, Qian, and Chen]{xing2021invertible}
Yazhou Xing, Zian Qian, and Qifeng Chen.
\newblock Invertible image signal processing.
\newblock In \emph{CVPR}, 2021.

\bibitem[Yang et~al.(2022)Yang, Jin, Xu, Zhang, Chen, and Liu]{yang2022seplut}
Canqian Yang, Meiguang Jin, Yi Xu, Rui Zhang, Ying Chen, and Huaida Liu.
\newblock {SepLUT}: Separable image-adaptive lookup tables for real-time image enhancement.
\newblock In \emph{ECCV}, 2022.

\bibitem[Zeng et~al.(2024)Zeng, Wang, Shen, and Wang]{zeng2024kan}
Chen Zeng, Jiahui Wang, Haoran Shen, and Qiao Wang.
\newblock Kan versus mlp on irregular or noisy functions.
\newblock \emph{arXiv preprint arXiv:2408.07906}, 2024.

\bibitem[Zhang et~al.(2023)Zhang, Tian, Li, Xu, Lu, Gao, and Sang]{zhang2023lookup}
Feng Zhang, Ming Tian, Zhiqiang Li, Bin Xu, Qingbo Lu, Changxin Gao, and Nong Sang.
\newblock Lookup table meets local laplacian filter: Pyramid reconstruction network for tone mapping.
\newblock In \emph{Thirty-seventh Conference on Neural Information Processing Systems}, 2023.

\bibitem[Zhang et~al.(2021)Zhang, Wang, Liu, Wang, Zhang, and Zuo]{zhang2021learning}
Zhilu Zhang, Haolin Wang, Ming Liu, Ruohao Wang, Jiawei Zhang, and Wangmeng Zuo.
\newblock Learning raw-to-s{RGB} mappings with inaccurately aligned supervision.
\newblock In \emph{ICCV}, 2021.

\bibitem[Zhou and Glotzbach(2007)]{zhou2007image}
Jianping Zhou and John Glotzbach.
\newblock Image pipeline tuning for digital cameras.
\newblock In \emph{IEEE International Symposium on Consumer Electronics}, 2007.

\bibitem[Zhu et~al.(2017)Zhu, Park, Isola, and Efros]{zhu2017unpaired}
Jun-Yan Zhu, Taesung Park, Phillip Isola, and Alexei~A Efros.
\newblock Unpaired image-to-image translation using cycle-consistent adversarial networks.
\newblock In \emph{Proceedings of the IEEE international conference on computer vision}, pages 2223--2232, 2017.

\end{thebibliography}
}

\end{document}